%% file: main.tex
\documentclass[10pt,twocolumn,letterpaper]{article}

\newcommand*{\FORARXIV}{}%

\ifdefined\FORREVIEW
\usepackage[review]{cvpr}      
\fi

\ifdefined\FORCAMERAREADY
\usepackage{cvpr}              
\fi

\ifdefined\FORARXIV
\usepackage[pagenumbers]{cvpr} 
\fi

\usepackage{graphicx}
\usepackage{amsmath}
\usepackage{amssymb}
\usepackage{booktabs}

\usepackage[pagebackref,breaklinks,colorlinks]{hyperref}

\usepackage[capitalize]{cleveref}
\crefname{section}{Sec.}{Secs.}
\Crefname{section}{Section}{Sections}
\Crefname{table}{Table}{Tables}
\crefname{table}{Tab.}{Tabs.}

\def\cvprPaperID{6269} 
\def\confName{CVPR}
\def\confYear{2022}

\input{preamble}
\begin{document}

\input{sec/0_metadata}
\maketitle
\input{sec/0_abstract}

\input{sec/1_introduction}
\input{sec/2_related}
\input{sec/3_method}
\input{sec/4_results}

\input{sec/5_conclusions}
\input{sec/6_acknowledgements}

{
    \small
    \bibliographystyle{ieee_fullname}
    \bibliography{macros,main}
}


\input{sec/X_supplementary}



\end{document}

%% file: preamble.tex
\usepackage{overpic}
\usepackage{enumitem} 
\usepackage{overpic} 
\usepackage{color}
\usepackage{colortbl} 
\usepackage{adjustbox} 
\usepackage{soul} 
\usepackage{dblfloatfix}
\usepackage{wrapfig}
\usepackage[T1]{fontenc}

\definecolor{teal}{rgb}{0,.5,.5}
\definecolor{ligh_teal}{rgb}{0.51,.79,.77}

\definecolor{blue}{rgb}{0,0,1}
\definecolor{red}{rgb}{1,0,0}
\definecolor{green}{rgb}{0,1.,0}
\definecolor{orange}{rgb}{0.75, 0.4, 0}
\definecolor{turquoise}{cmyk}{0.65,0,0.1,0.3}
\definecolor{purple}{rgb}{0.65,0,0.65}
\definecolor{darkjunglegreen}{rgb}{0.1, 0.14, 0.13}
\definecolor{darkgreen}{rgb}{0.0, 0.2, 0.13}
\definecolor{darkred}{rgb}{0.6, 0.1, 0.05}
\definecolor{blueish}{rgb}{0.0, 0.3, .6}
\definecolor{light_gray}{rgb}{0.7, 0.7, .7}
\definecolor{pink}{rgb}{1, 0, 1}
\definecolor{greyblue}{rgb}{0.25, 0.25, 1}
\definecolor{magenta}{rgb}{1., 0., 1.}

\newcommand{\shortcite}[1]{\cite{#1}}

\usepackage{blindtext}

\renewcommand{\paragraph}[1]{\vspace{1em}\noindent\textbf{#1}.}

%% file: sec/0_metadata.tex
\title{ShapeFormer: Transformer-based Shape Completion via Sparse Representation
}
\author{
Xingguang Yan$^{1}$\hspace{.01cm}
Liqiang Lin$^{1}$\hspace{.01cm}
Niloy J. Mitra$^{2,3}$\hspace{.01cm}
Dani Lischinski$^{4}$\hspace{.01cm}
\ifdefined\FORARXIV
Daniel Cohen-Or$^{5}$\hspace{.01cm}
\fi
\ifdefined\FORCAMERAREADY
Daniel Cohen-Or$^{1,5}$\hspace{.01cm}
\fi
Hui Huang$^{1*}$\\
\small
$^{1}$Shenzhen University\hspace{.02cm}
$^{2}$University College London\hspace{.02cm}
$^{3}$Adobe Research\hspace{.02cm}
$^{4}$Hebrew University of Jerusalem\hspace{.02cm}
$^{5}$Tel Aviv University
 \\
}

%% file: sec/0_abstract.tex
\begin{abstract}

We present \textit{ShapeFormer}, a transformer-based network that produces a distribution of object completions, conditioned on incomplete, and possibly noisy,  point clouds. The resultant distribution can then be sampled to generate likely completions, each exhibiting plausible shape details while being faithful to the input. 

To facilitate the use of transformers for 3D, we introduce a compact 3D representation, \textit{vector quantized deep implicit function} (VQDIF), that utilizes spatial sparsity to represent a close approximation of a 3D shape by a  short sequence of discrete variables. Experiments demonstrate that ShapeFormer outperforms prior art for shape completion from ambiguous partial inputs in terms of both completion quality and diversity. We also show that our approach effectively handles a variety of shape types, incomplete patterns, and real-world scans.

\end{abstract}

%% file: sec/1_introduction.tex
\section{Introduction}
\label{sec:intro}
\input{fig/teaser}

\let\thefootnote\relax\footnote{* Corresponding author: Hui Huang (hhzhiyan@gmail.com)} 
\let\thefootnote\relax\footnote{Project page: \href{https://shapeformer.github.io}{https://shapeformer.github.io}
} 

Shapes are typically acquired with cameras that probe and sample surfaces. The process relies on line of sight and, at best, can obtain partial information from the visible parts of objects. Hence, sampling complex real-world geometry is inevitably imperfect, resulting in varying sampling densities and missing parts. This problem of surface completion has been extensively investigated over multiple  decades~\cite{berger2017survey}. The central challenge is to compensate for incomplete data by inspecting non-local hints in the observed data to infer missing parts using various forms of priors. 

Recently, deep implicit function~(DIF) has
emerged as an effective  representation for learning high-quality surface completion. To learn shape priors, earlier DIFs~\cite{zhiqin2019imnet, lars2019occnet, park2019deepsdf} encode each shape using a single global latent vector. Combining a global code with region-specific local latent codes~\cite{Erler2020Points2Surf, Peng2020ConvONet, chibane20ifnet, jiang2020lig, genova2020ldif, chen2021mdif} can faithfully preserve geometric details of the input in the completion.
However, when presented with \emph{ambiguous} partial input, for which multiple plausible completions are possible (see \cref{fig:teaser}), the deterministic nature of local DIF usually fails to produce meaningful completions for unseen regions.
A viable alternative is to combine generative models to handle the input uncertainty.
However, for representations that contain enormous statistical redundancy, as in the case of current local methods, such combination \cite{sun2020pointgrow} 
excessively allocates model capacity towards perceptually irrelevant details \cite{Fauw2019HierarchicalAR, dieleman2021SlowAEs}. 

We present \textit{ShapeFormer}, a transformer-based autoregressive model that learns a \textit{distribution} over possible shape completions. We use local codes to form a sequence of discrete, vector quantized features, greatly reducing the representation size while keeping the underlying structure. Applying transformer-based generative models toward such sequences of discrete variables have been shown to be effective for generative pretraining~\cite{chen2020imagegpt, bao2021beit}, generation~\cite{razavi2019vqvae2,esser2020taming} and completion~\cite{wan2021ict} in image domain.

However, directly deploying transformers to 3D feature grids leads to a sequence length cubic in the feature resolution. Since transformers have an innate quadratic complexity on sequence length, only using overly coarse feature resolution, while feasible, can barely represent meaningful shapes.
To mitigate the complexity, we first introduce Vector Quantized Deep Implicit Functions~(\textit{VQDIF}), a novel 3D representation that is both compact and structured, that can represent complex 3D shapes with acceptable accuracy, while being rather small in size.
The core idea is to sparsely encode shapes as sequences of discrete 2-tuples, each representing both the position and content of a non-empty local feature. 
These sequences can be decoded to deep implicit functions from which high-quality surfaces can subsequently be extracted.
Due to the sparse nature of 3D shapes, such encoding reduces the sequence length from cubic to quadratic in the feature resolution, thus enabling effective combination with generative models.

\textit{ShapeFormer} completes shapes by generating complete sequences, conditioned on the sequence for partial observation.
It is trained by sequentially predicting the conditional distribution of both location and content over the next element.
Unlike image completion~\cite{wan2021ict}, where the model is trained with the BERT\cite{devlin2018bert,bao2021beit} objective to only predict for unseen regions, in the 3D shape completion setting, the input features may also come from both noisy and incomplete observations, and keeping them intact necessarily yields noisy results. Hence, in order to generate whole complete sequences from scratch while being faithful to the partial observations, we adapt the auto-regressive objective and prepend the partial sequence to the complete one to achieve conditioning. This strategy has been proved effective for conditional synthesis for both text~\cite{liu2018wikigen} and images~\cite{esser2020taming}.

We demonstrate the ability of ShapeFormer to produce diverse high-quality completions for ambiguous partial observations of various shape types, including CAD models and human bodies, and of various incomplete sources such as real-world scans with missing parts. 
In summary, our contributions include: (i)~a novel DIF representation based on sequences of discrete variables that compactly represents satisfactory approximations of 3D shapes; (ii)~a transformer-based autoregressive model that uses our new representation to predict multiple high-quality completed shapes conditioned on the partial input; and (iii)~state-of-the-art results for multi-modal shape completion in terms of completion quality and diversity. The FPD score on PartNet is improved by at most 1.7 compared with prior multi-modal method cGAN~\cite{wu2020cGAN}.

%% file: fig/teaser.tex

\begin{figure}[!t]
	\includegraphics[width=\linewidth]{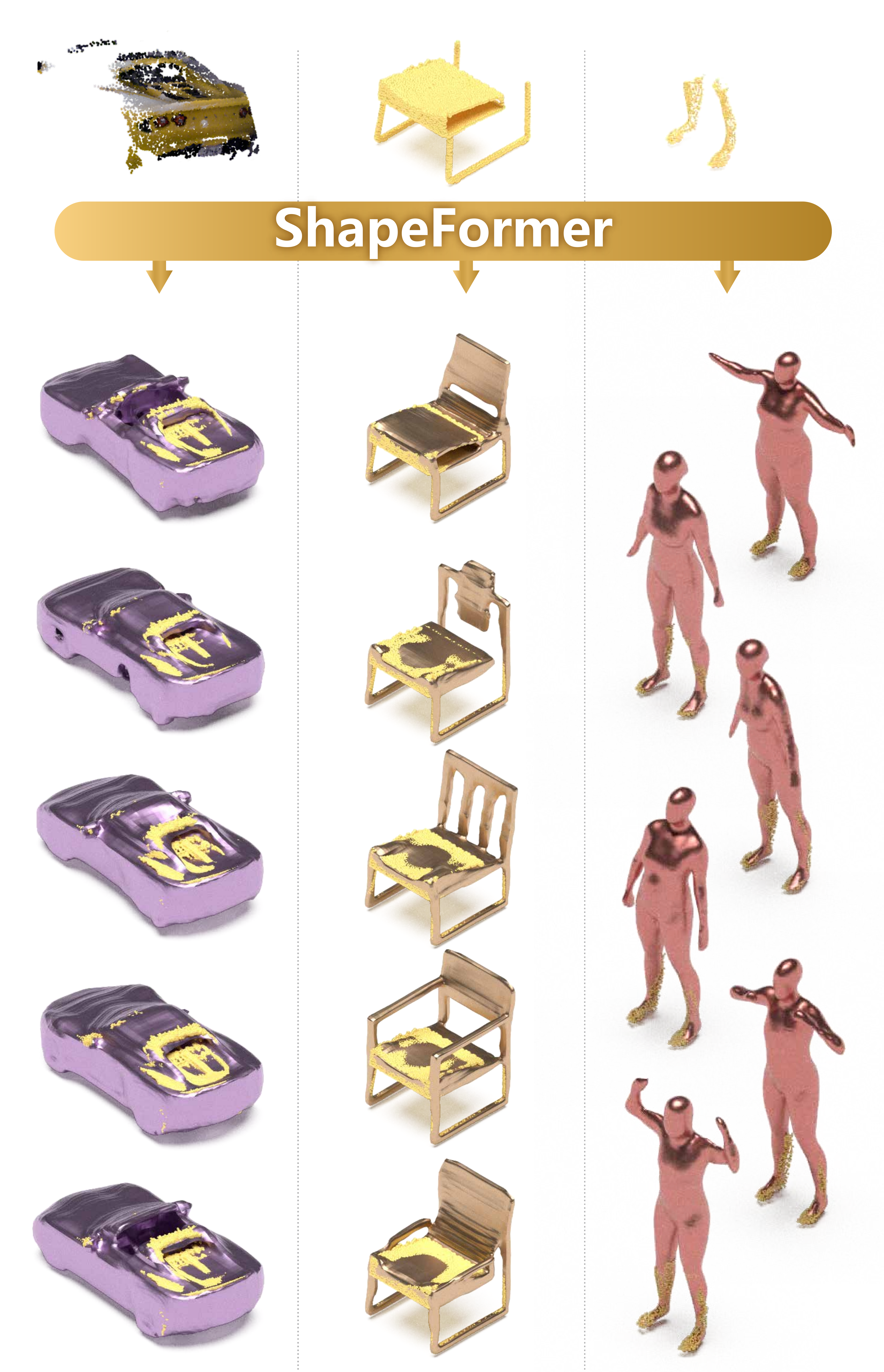}
	\caption{ 
	ShapeFormer predicts multiple completions for a real-world scan of a sports car (left column), a chair with missing parts (middle column), and a partial point cloud of human lower legs (right column). The input point clouds are superimposed with the generated shapes to emphasize the faithfulness of the completion to the input point cloud.
	}
	\label{fig:teaser}
\end{figure}

%


%% file: sec/2_related.tex
\section{Related Work}
\label{sec:related}
\input{fig/pipeline_sampling}
\paragraph{Shape reconstruction and completion}
3D reconstruction is a longstanding ill-posed problem in computer vision and graphics. Traditional methods can produce faithful reconstruction from complete input such as point cloud~\cite{berger2017survey}, or images~\cite{mvstutorial}. Recently, neural network-based methods have demonstrated an impressive performance toward reconstruction from partial input~\cite{han2019image}, where the unseen regions are completed with the help of data priors. They can be classified according to their output representation, such as voxels, meshes, point clouds, and deep implicit functions. 
Since voxels can be processed or generated easily through 3D convolutions thanks to their regularity, they are commonly used in earlier works~\cite{dai20173depn, stutz2018completion, choy20163dr2n2,Hne2017hsp}. However, since their cubic complexity toward resolution, the predicted shapes are either too coarse or too heavy in size for later applications.
While meshes are more data-efficient, due to the difficulty of handling mesh topology, mesh-based methods have to either use shape template~\cite{rock2015completing,wang2018pixel2mesh,litany2018deformable}, limiting to a single topology, or produce self-intersecting meshes~\cite{groueix2018atlasnet}.
Point clouds, in contrast, do not have such a problem and are popularly used lately for generation~\cite{fan2017pointgen, Achlioptas2017pointae} and completion~\cite{yuan2018pcn, Tchapmi2019topnet, yu2021pointr,Xiang2021SnowflakeNet}.
However, point clouds need to be non-trivially post-processed using
classical methods~\cite{bernardini1999ball, kazhdan2006poisson,kazhdan2013poisson,huang2019VIPSS} to recover surfaces due to their sparse nature.
Recent works that represent shapes as deep implicit functions have been shown to be effective for high-quality 3D reconstruction~\cite{zhiqin2019imnet,lars2019occnet,park2019deepsdf}. 
By leveraging local priors, follow-up works~\cite{chibane20ifnet,Peng2020ConvONet,Erler2020Points2Surf,liu2021IMLSNet,genova2020ldif} can further improve the fidelity of geometric details. 
However, most current methods are not effective toward ambiguous input due to their deterministic nature.
Other methods handle such input by leveraging generative models. They learn the conditional distribution of complete shapes represented as either a single global code~\cite{wu2020cGAN,arora2021multimodal}, which, due to their lack of spatial structure, leads to completions misaligned with the input, or raw point cloud~\cite{zhou2021pvd}, which, due to its statistical redundancy, is only effective for completing simple shapes with a limited number of points.
In this paper, we show how building generative models upon our new compact, structured representation enables multi-modal high-quality reconstruction for complex shapes.

\paragraph{Autoregressive models and Transformers}
Autoregressive models are generative models that aim to model distributions of high dimensional data by factoring the joint probability distribution to a series of conditional distributions via the chain rule~\cite{bengio2000modeling}.
Using neural networks to parameterize the conditional distribution has been proved to be effective~\cite{uria2016nade,germain2015made} in general, and more specifically to image generation~\cite{van2016pixelrnn,oord2016pixelcnn,chen2018pixelsnail}.
Transformers~\cite{vaswani2017transformer}, known for their ability to model long-range dependencies through self-attentions, have shown the power of autoregressive models in natural languages~\cite{radford2019GPT2,brown2020GPT3}, image generation~\cite{parmar2018imagetransformer,chen2020imagegpt}. 
Contrary to deterministic masked auto-encoders~\cite{he2021mae}, Transformers can produce diverse image completions~\cite{wan2021ict} that are sharp in masked regions by adopting the BERT~\cite{devlin2018bert} training objective. 
In the 3D domain, autoregressive models have been used to learn the distribution of point clouds~\cite{sun2020pointgrow,wang2020sceneformer} and meshes~\cite{nash2020polygen}. 
However, these models can only generate small point clouds or meshes restricted to 1024 vertices due to the lack of efficient representation. 
In contrast, by eliminating statistical redundancy, a compressed discrete representation enables generative models to focus on data dependencies at a more salient level~\cite{van2017vqvae1,razavi2019vqvae2} and recently allows high-resolution image synthesis~\cite{esser2020taming,ramesh2021dalle}. Follow-up works utilize data sparsity to obtain even more compact representations~\cite{nash2021DCTransformer,dieleman2021SlowAEs}. 
We explore this direction in the context of surface completion.
Concurrently with our work, AutoSDF~\cite{Mittal2022AutoSDF} trains Transformers to complete and generation shapes with dense grid. And Point-BERT~\cite{Yu2021PointBERT} adopts generative pre-training for several downstream tasks.

%% file: fig/pipeline_sampling.tex
\begin{figure*}[!t]
    \centering
    \includegraphics[width=\textwidth]{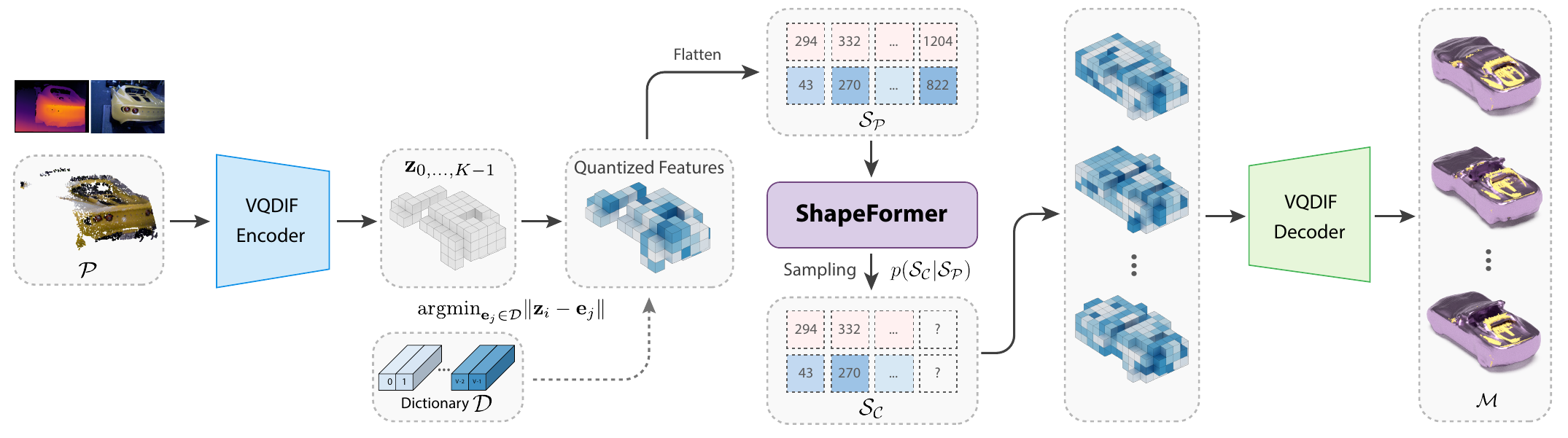}
\caption{Overview of our shape completion approach. Given a partial point cloud $\mathcal{P}$, possibly from a depth image, as input, our VQDIF encoder first converts it to a sparse feature sequence $\mathbf{z}_{0...K-1}$, replacing them with the indices of their nearest neighbor $\mathbf{e}_j$ in a learned dictionary $\mathcal{D}$, forming a sequence of discrete 2-tuples consisting of the coordinate (pink) and the quantized feature index (blue). 
We refer to this partial sequence as $\mathcal{S}_\mathcal{P}$ (drawn with dashed lines).
The ShapeFormer then takes  $\mathcal{S}_\mathcal{P}$ as input and models the conditional distribution $p(\mathcal{S}_\mathcal{C} | \mathcal{S}_\mathcal{P})$.
Autoregressive sampling yields a probable complete sequence $\mathcal{S}_\mathcal{C}$. Finally, the VQDIF decoder converts the sequence $\mathcal{S}_\mathcal{C}$ to a deep implicit function, from which the surface reconstruction $\mathcal{M}$ can be extracted. To show the faithfulness of our reconstructions, we super-impose the input point cloud on them.
Please see the supplementary material for more architectural details.
}

\label{fig:pipeline_sampling} 
\end{figure*}

%% file: sec/3_method.tex
\section{Method}

We model the shape completion problem as mapping a partial point cloud $\mathcal{P}\in\mathbb{R}^{N\times 3}$ to a complete, watertight mesh $\mathcal{M}$ which matches the cloud. Since this is an ill-posed problem, we seek to estimate the probabilistic distribution of such mesh $p(\mathcal{M}|\mathcal{P})$ utilizing the power of Transformers.
Instead of working directly on point clouds, meshes, or feature grids, we approximate shapes as short discrete sequences (see \cref{sec:VQDIF}) to greatly reduce both the number of variables and the variable bit size, which enables Transformers to complete complex 3D shapes (see \cref{sec:ShapeFormer}).

With such compact representation, the conditional distribution becomes $p(\mathcal{S}_\mathcal{C}|\mathcal{S}_\mathcal{P})$, where $\mathcal{S}_\mathcal{P}$ and $\mathcal{S}_\mathcal{C}$ are the sequence encoding of the partial point cloud and the complete shape, respectively.
Once such distribution is modeled, we can sample multiple complete sequences $\mathcal{S}_\mathcal{C}$, from which different surface reconstructions $\mathcal{M}$ can be obtained through decoding. This process is illustrated in \cref{fig:pipeline_sampling}.

\subsection{Compact sequence encoding for 3D shapes} \label{sec:VQDIF}
We propose VQDIF, whose goal is to approximate 3D shapes with a shape dictionary, with each entry describing a particular type of local shape part inside a cell of volumetric grid $G$ with resolution $R$.
With such a dictionary, shapes can be encoded as short sequences of entry indices, describing the local shapes inside all non-empty grid cells, enabling transformers to efficiently model the global dependencies.

We design an auto-encoder architecture to achieve this. The encoder $E$ first maps the input point cloud to a 64 resolution feature grid with local-pooled PointNet and then downsample it to resolution $R$. Unlike the previous strategy for image synthesis~\cite{esser2020taming}, the encoder parameters are carefully set to have the least receptive field, reducing the number of non-empty features to the number of sparse voxels of the voxelized input point cloud $\mathcal{P}$ at resolution $R$.
Then these non-empty features are flattened to a sequence of length $K$ in row-major order. Since these features are sparse, we record their locations with their flattened index $\{c_i\}_{i=0}^{K-1}$.
Other orderings are also possible, but for generation they are not as effective as row-major order~\cite{esser2020taming}.

Following the idea of neural discrete representation learning \cite{van2017vqvae1}, we compress the bit size of the feature sequence $\{\mathbf{z}_i\}_{i=0}^{K-1}$ through vector quantization, that is, clamping it to its nearest entry in a dictionary $\mathcal{D}$ of $V$ embeddings $\{\mathbf{e}_j\}_{j=0}^{V}$ and we save the indices of these entries:
\begin{equation}
    v_i = \text{argmin}_{j\in[0,V)} \|\mathbf{z}_i-\mathbf{e}_j\|.
\end{equation}
Thus, we get a compact sequence of discrete 2-tuples representing the 3D shape $\mathcal{S}=\{(c_i, v_i)\}_{i=0}^{K-1}$.
Finally, the decoder projects this sequence back to a feature grid and, through a 3D-Unet~\cite{cciccek2016unet3d}, decodes it to a local deep implicit function $f$~\cite{Peng2020ConvONet}, whose iso-surface is the reconstruction $\mathcal{M}$.

\paragraph{Training} We train the VQDIF by simultaneously minimizing the reconstruction loss and updating the dictionary using exponential moving averages~\cite{van2017vqvae1}, where dictionary embeddings are gradually pulled toward the encoded features. 
We also adopt commitment loss $\mathcal{L}_{\text{commit}}$~\cite{van2017vqvae1} to encourage encoded features $\mathbf{z}_i$ to stay close to their nearest entry $\mathbf{e}_{v_i}$ in the dictionary, with index $v_i$, thus keeping the range of the embeddings bounded. We define the loss as, 
\begin{equation}
    \mathcal{L}_\text{commit} = \frac{1}{K}\sum_{i=0}^{K-1}(\mathbf{z}_i-\text{sg}[\mathbf{e}_{v_i}])^2, 
\end{equation}
where sg stands for stop gradient operator which prevents the embedding being affected by this loss.

The full training objective for VQDIF is the combination of reconstruction loss of $\mathcal{L}_\text{commit}$ with weighting factor $\beta$:
\begin{equation}
    \mathcal{L}_\text{\tiny{VQDIF}} = \frac{1}{T}\sum_{i=0}^{T-1}\text{BCE}\large( f(\mathbf{x}_i), o_i \large) 
                           + \beta  \mathcal{L}_\text{commit}.
\end{equation}
Here, $T$ is the size of the target set and BCE is the binary cross-entropy loss which measures the discrepancy between the predicted and the ground truth occupancy $o_i$ at target point $\mathbf{x}_i$. During training, we select the target set $\mathcal{T}_x=\{\mathbf{x}_{i=0}^{T-1}\}$ and its occupancy values $\mathcal{T}_o=\{o_{i=0}^{T-1}\}$ in a similar fashion to prior work~\cite{lars2019occnet}.

\input{fig/pipeline_shapeformer}
\subsection{Sequence generation for shape completion} \label{sec:ShapeFormer}
\input{fig/cmp_implicit}
We autoregressively model the distribution $p(\mathcal{S}_\mathcal{C}|\mathcal{S}_\mathcal{P})$, by predicting the distribution of the next element conditioned on the previous elements. We also factor out the tuple distribution for each element: $p(c_i,v_i)=p(c_i)  p(v_i | c_i)$ . The final factored sequence distribution is as follows:
\begin{align*} 
p(\mathcal{S}_\mathcal{C}|\mathcal{S}_\mathcal{P}; \theta)
        &= \prod_{i=0}^{K-1}   p_{c_i}\cdot p_{v_i} \\
p_{c_i} &= p(c_i|\mathbf{c}_{   < i}, \mathbf{v}_{<i}, \mathcal{S}_\mathcal{P}; \theta) \\ 
p_{v_i} &= p(v_i|\mathbf{c}_{\leq i}, \mathbf{v}_{<i}, \mathcal{S}_\mathcal{P}; \theta).
\label{sf_prob}
\end{align*}
Here, $\theta$ indicates model parameters and $p_{c_i}$ and $p_{v_i}$ are the distributions of the coordinate and the index value of the $i$-th element of $\mathcal{S}_\mathcal{C}$, conditioned on previously generated elements and the partial sequence $\mathcal{S}_\mathcal{P}$. Note that $p_{v_i}$ is also conditioned on the current coordinate $c_i$.

Different approaches have been applied to build a transformer model that can predict tuple sequences. 
Instead of flattening them \cite{sun2020pointgrow}, which in our case doubles the sequence length, we stack two decoder-only transformers to predict the $p_{c_i}$ and $p_{v_i}$ respectively in a similar way to prior works \cite{wang2020sceneformer, nash2021DCTransformer, dieleman2021SlowAEs}, as illustrated in \cref{fig:pipeline_shapeformer}.
Unlike in the image completion case~\cite{wan2021ict}, where the partial sequence is strictly a part of the complete sequence so that only the missing regions need to be completed. For our case, however, due to the noise or incompleteness of local observations, we would like to predict complete sequences from scratch to fix such data deficiencies.
And thanks to the autoregressive structure of the decoder-only transformer, we can achieve conditioning by simply prepending $\mathcal{S}_\mathcal{P}$ before $\mathcal{S}_\mathcal{C}$ to generate complete sequences that are in coordination with the partial one.
We also append an additional end token to both sequences to help learning.

\paragraph{Training and inference}
The training objective of ShapeFormer is to maximize the log-likelihood given both $\mathcal{S}_\mathcal{C}$ and $\mathcal{S}_\mathcal{P}$:
 $   \mathcal{L}_{\text{ShapeFormer}}=-\log p(\mathcal{S}_\mathcal{C}|\mathcal{S}_\mathcal{P}; \theta).$
After the model is trained, ShapeFormer performs shape completion by sequentially sampling the next element of the complete sequence until an end token ([END]) is encountered. Given the partial sequence, we alternatively sample the new coordinate and value index using top-p sampling \cite{holtzman2019nucleus}, where only a few top choices, for which the sum of probabilities exceeds a threshold $p_n$, are kept.
Also, we mask out the invalid choices for coordinate to guarantee monotonicity.

%% file: fig/pipeline_shapeformer.tex
\begin{figure}[!t]
	\includegraphics[width=\linewidth]{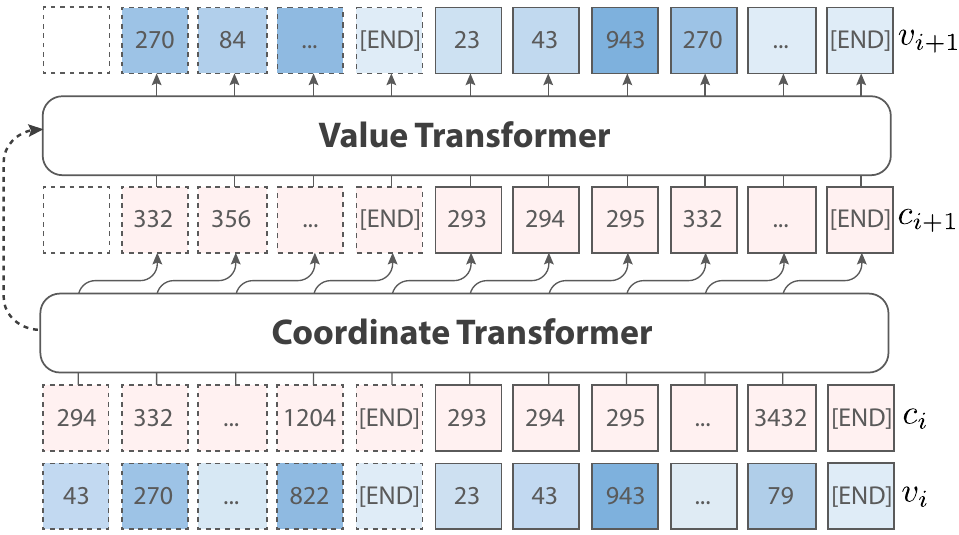}
	\caption{
	The architecture of the ShapeFormer. 
	The partial sequence $\mathcal{S}_\mathcal{P}$ (dashed boxes) and the complete one $\mathcal{S}_\mathcal{C}$ (solid boxes) both appended with an end token are concatenated before sending their locations, $c_i$ (pink) and values $v_i$ (blue), to a Coordinate Transformer to predict the next location $c_{i+1}$. The Value Transformer takes both $c_{i+1}$ and the former Transformer's output embedding to predict the next value $v_{i+1}$.
}
	\label{fig:pipeline_shapeformer}
\end{figure}


%% file: fig/cmp_implicit.tex
\begin{figure*}[!t]
    \centering
    \includegraphics[width=\textwidth]{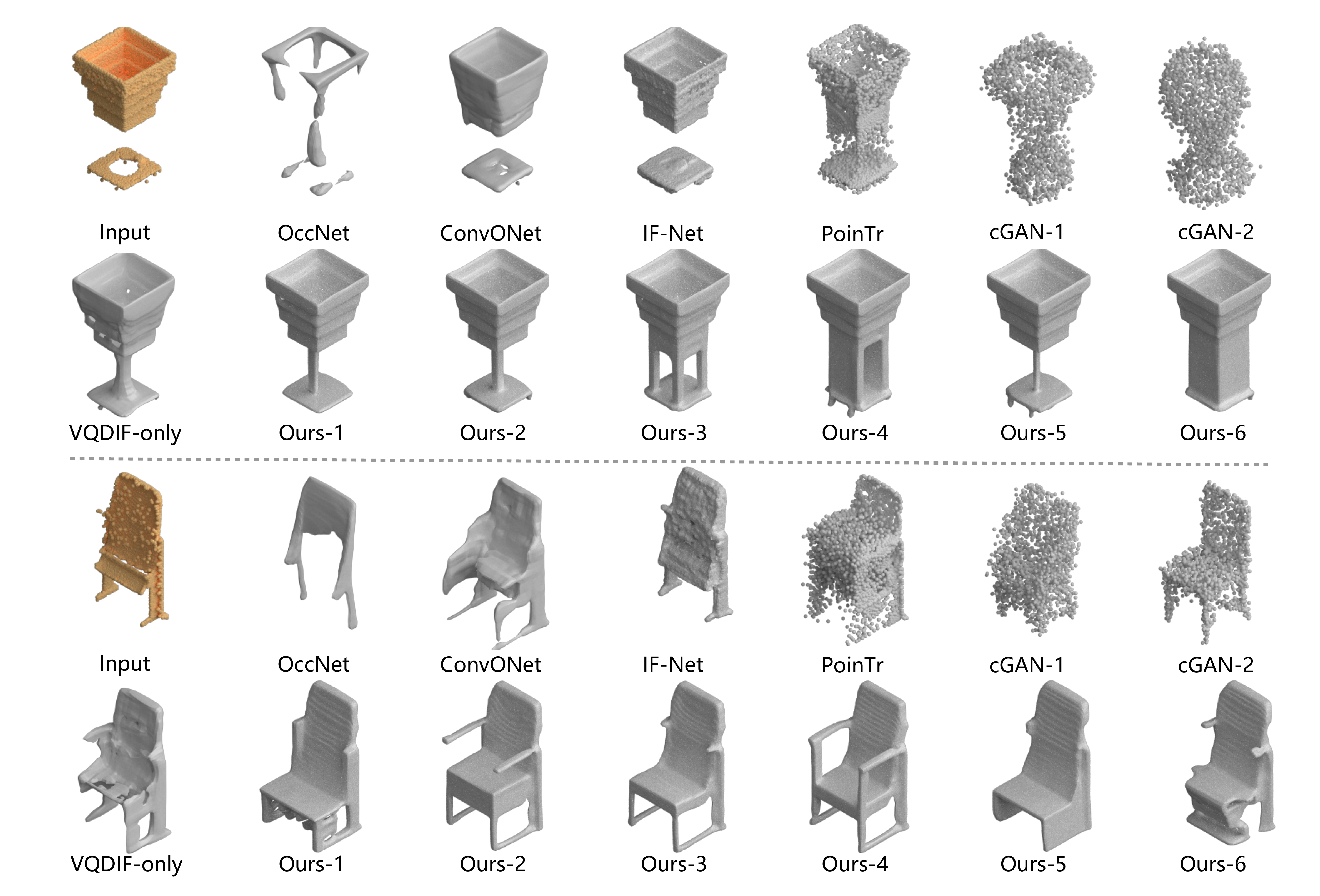}
    \caption{Visual comparison with prior shape completion methods on the ShapeNet dataset.
    Our method can better handle ambiguous scans and produce completions that are more faithful on both observed and unseen regions.
    More examples are in the supplementary material.
}
    \label{fig:cmp_implicit} 
\end{figure*}

    

%% file: sec/4_results.tex
\section{Results and Evaluation}
\input{fig/cmp_partnet}
\input{tab/table_cmp_hprscan}

In this section, we demonstrate our method outperforms prior arts for shape completion from ambiguous scans and part-level incompleteness (\cref{sec:cmp_shape_completion}).
Then we show our approach can effectively handle a variety of shape types, 
out-of-distribution shapes, and real-world scans from the Redwood dataset~\cite{choi2016redwood} (\cref{sec:more_results}).
Lastly, we show our VQDIF representation has a significantly smaller size compared with prior DIFs while achieving similar accuracy (\cref{sec:cmp_shape_recon}).

Throughout all these experiments, we use feature resolution $R=16$ for VQDIF and set its loss balancing factor $\beta=0.01$. We also set the vocabulary of the dictionary $\mathcal{D}$ to be $V=4096$. We use $20$ and $4$ blocks for Coordinate and Value Transformers, respectively. 
All of these blocks have $16$ heads self-attention, and the embedding dimension is $1024$.
We find that a maximum sequence length of $812$ is enough for all of our experiments. 
We set the default probability factor $p=0.4$ for sampling.
Further implementation details such as architecture and training statistics are provided in the supplementary.

\subsection{Shape completion results} \label{sec:cmp_shape_completion}
\paragraph{Data}
We consider two datasets: 1) ShapeNet~\cite{chang2015shapenet} for testing on partial scan and 2) PartNet~\cite{Mo2019PartNet} for testing on part-level incompleteness; we follow the same setting as in cGAN~\cite{wu2020cGAN}.
For ShapeNet, following prior works~\cite{lars2019occnet, chibane20ifnet, Peng2020ConvONet, zhiqin2019imnet}, we use 13 classes of the ShapeNet with train/val/test split from 3D-R2N2~\cite{choy20163dr2n2}. The data are processed and sampled similarly to IMNet~\cite{zhiqin2019imnet} and we create partial input for training via random virtual scanning. 
For evaluation, we first measure the ambiguity score of a partial point cloud $\mathcal{P}$ to its complete counterpart $\mathcal{C}$ as the mean ratio of the distance of each point $x\in\mathcal{C}$ with its nearest neighbor in $\mathcal{P}$ to its distance toward furthest neighbor in $\mathcal{C}$. We uniformly sample 70 viewpoints on a sphere for each shape. Then we create two setups for the dataset according to ambiguity. The high scan ambiguity setup selects scans with the top half ambiguity score and vice versa. More details about this score are provided in the supplementary material.

\input{tab/table_partnet}
\paragraph{Metric}
For the low ambiguity setting, we use Chamfer $L_2$ Distance (\textit{CD}) and F-score\%1 (\textit{F1})~\cite{tatarchenko2019single} to measure how accurate the completion is; this is similar to the previous setup~\cite{Peng2020ConvONet}.
And to evaluate completion quality for high ambiguity setting, we follow prior work~\cite{shu2019treegan} to use pre-trained PointNet~\cite{qi2017pointnet} classifier as a feature extractor to compute the Fr\'echet Point Cloud Distance (FPD) between the set of completion results and ground truth shapes.
Additionally, for the PartNet dataset, we follow cGAN~\cite{wu2020cGAN} and use Unidirectional Hausdorff Distance (UHD) to measure faithfulness toward input, Total Mutual Difference (TMD) to measure diversity, and Minimal Matching Distance (MMD)\cite{Achlioptas2017pointae}.

\paragraph{Baselines}
We compare our model with a global DIF method OccNet~\cite{lars2019occnet}, two local DIF methods ConvONet~\cite{Peng2020ConvONet} and IF-Net~\cite{chibane20ifnet}, PoinTr~\cite{yu2021pointr}, which adopts Transformers without autoregressive learning, and multi-modal completion method cGAN~\cite{wu2020cGAN}. 
We also compare our VQDIF-only model to illustrate the necessity of ShapeFormer.
We train these methods for shape completion in our dataset setting with their official implementation.

\paragraph{Results on ShapeNet}
As shown in \cref{fig:cmp_implicit}, methods incorporating structured local features can better preserve the input details than those that only operate on global features (OccNet~\cite{lars2019occnet}, cGAN~\cite{wu2020cGAN})
And deterministic methods tend to produce averaged shape since they are unable to handle multi-modality. 
Notice that PoinTr~\cite{yu2021pointr} also utilizes the power of Transformers, but they can not alleviate this problem by adopting Transformers without generative modeling.
This phenomenon is more apparent for the chair example, which has higher ambiguity. Our VQDIF-only model also fails to produce good completion in this case.
Based on VQDIF, our ShapeFormer resolves ambiguity by factoring the estimation into a distribution, with each sampled shape sharp and plausible.
In contrast, the multi-modal method cGAN~\cite{wu2020cGAN} is unable to produce high-quality shapes due to their unstructured representation.
Further, we generate one completion per input with top-p sampling for quantitative evaluation.
As shown in \cref{tab:cmp_implicit}, our method has a much better FPD for high ambiguity scans.
Notice CD is not reliable when ambiguity is high since it often treats plausible completions as significant errors. For low ambiguity scans, our method is also competitive toward previous state-of-the-art completion methods in terms of accuracy.



\input{fig/results_realscan}
\input{fig/results_famous}

\paragraph{Results on PartNet}
We compare our model with cGAN and ShapeInversion~\cite{zhang2021shapeinversion} on PartNet. The latter method achieves multiple completions through GAN inversion. The quantitative and qualitative comparisons are shown in \cref{tab:cmp_partnet} and \cref{fig:cmp_partnet}, respectively. Thanks to our structured representation, we achieve much better faithfulness (UHD) and can generate more varied (TMD) high-quality shapes (MMD and FPD) than these GAN-based methods.

\input{fig/results_human}
\subsection{More results} \label{sec:more_results}

\paragraph{Results on real scans}
We further investigate how our model pre-trained on ShapeNet can be applied to scans of real objects. We test our model on partial point clouds converted from RGBD scans of the Redwood 3D Scans dataset~\cite{choi2016redwood}.
\Cref{fig:results_realscan} shows the results for a sofa and a table, both of them have two scans from different views.
Notice that our model sensitively captures the uncertainty of a scan, producing a distribution of completions that are faithful to the scan and plausible in unobserved regions.
We also show results for a sports car in \cref{fig:pipeline_sampling}.

\paragraph{Results on out-of-distribution objects}
We further evaluate ShapeFormer's generalization by testing scans of unseen types of shapes on our trained model of \cref{sec:cmp_shape_completion}.
We pick the novel shapes from the "Famous" dataset collected by Erler et al.~\shortcite{Erler2020Points2Surf} which includes many famous geometries for testing, such as the "Utah teapot," and apply virtual scan to get the partial point cloud.
\cref{fig:results_famous} demonstrates our ShapeFormer can grasp general concepts such as symmetry or hollow and filled.
Even the model is only trained on the 13 ShapeNet categories, without ever seeing any cups or teapot, it can still successfully produce multiple reasonable completions from the partial scan.
Moreover, in the second row, we see the completions of a one-side scan of a cup contain two distinct features: the cups might be solid or empty.
These examples show the ShapeFormer's potential for general-purpose shape completion, where once we have it trained, we can apply it for all types of shapes.

\input{tab/table_cmp_recon}

\input{fig/results_recon_bytes}

\paragraph{Results on human shapes}
    In addition to CAD models, we qualitatively evaluate our completion results on scans of human shapes (D-FAUST dataset~\cite{federica2017CVPRdfaust}) using the same setting as Niemeyer et al.~\cite{Niemeyer2019oflow}.
    Human shapes are very challenging due to their thin structures and the wide variety of poses.
    To simulate part level incompleteness, we randomly select a point from the complete cloud and only keep neighboring points within a ball of a fixed radius as partial input.
    \cref{fig:results_human} shows examples of our results.
    We can see that our completions keep the pose of the observed body parts and generate various possible poses for the unobserved body parts.
    
\subsection{Surface reconstruction with VQDIF} \label{sec:cmp_shape_recon}
    Our final experiment evaluates the representation size and reconstruction accuracy of VQDIF. 
    We compare VQDIF of different feature resolutions ($\text{Ours}_8$, $\text{Ours}_{16}$, $\text{Ours}_{32}$) with OccNet, ConvONet, IF-Net, which are retrained to auto-encode the complete shape with their released implementations.
    As shown in \cref{fig:results_recon_bytes}, $\text{Ours}_{32}$ achieves similar accuracy to the local implicit approach IF-Net while being significantly smaller in size thanks to the sparse and discrete VQDIF features.
    The minimum receptive field of our encoder keeps the feature as local as possible, which greatly reduces the feature amount.
    Then the multi-dimensional feature vectors are quantized and can be referred to using a single integer index, which further reduces the size. The accuracy loss is only salient for lower feature resolution, as seen in the \textit{w/o quant.} comparison, where we train VQDIF without vector quantization.
    These together allow transformers to effectively model the distribution of shapes.
    We adopt $\text{Ours}_{16}$ for ShapeFormer since it only has an average length of 217 (see \cref{tab:cmp_recon}) and its accuracy is already comparable with ConvONet (see \cref{fig:results_recon_cmp}).
\input{fig/results_recon_cmp}

%% file: fig/cmp_partnet.tex
\begin{figure*}[!t]
    \centering
    \includegraphics[width=\linewidth]{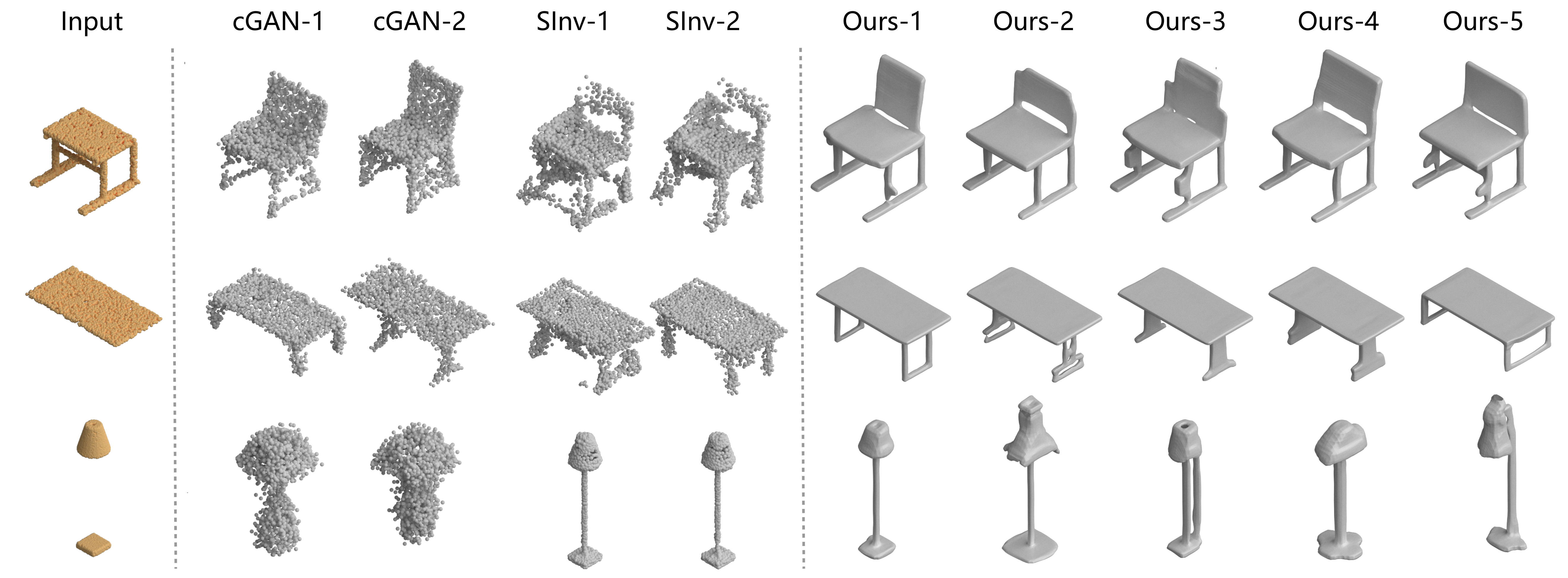}
    \caption{Visual comparison for multi-modal shape completion of Table, Chair, and Lamp categories on PartNet. 
        We can produce diverse completions that better align with the input.
    }
    \label{fig:cmp_partnet} 
\end{figure*}

%% file: tab/table_cmp_hprscan.tex
\begin{table}[t!]%
		\begin{adjustbox}{width=\columnwidth,center}
		\begin{tabular}{lccc|ccc}
        \toprule
          \textsc{Scan Ambiguity} & \multicolumn{3}{c}{\textsc{Low}} & \multicolumn{3}{c}{\textsc{High}}\\ 
         \midrule

          Method & CD$\downarrow$ & F1$\uparrow$ & FPD$\downarrow$& CD$\downarrow$ & F1$\uparrow$ & FPD$\downarrow$ \\ \hline    
            OccNet\cite{lars2019occnet}         & 1.48 & 63.2 & 0.34            & \textbf{2.79} & 50.4 & 3.12  \\ 
            ConvONet\cite{Peng2020ConvONet}     & 0.81 & 72.9           & 0.23  & 3.14 & 60.4 & 2.85  \\ 
            IF-Net\cite{chibane20ifnet}         & 0.79 & \textbf{73.8}  & 0.25  & 18.4 & 51.5 & 3.66  \\ 
            PoinTr\cite{yu2021pointr}           & 0.80 & 70.1 & 0.23            & 3.11 & 59.3 & 3.29  \\ 
            cGAN\cite{wu2020cGAN}               & 1.33 & 62.1 & 1.36            & 3.49  & 59.3& 2.55   \\ \hline
            Ours                                & 0.74 & 70.3 & 0.24            & 4.72 & 60.5 & \textbf{1.45}  \\ 
            $\text{Ours}^*$                     & \textbf{0.73} & 71.4 & \textbf{0.22} & 4.69 & \textbf{60.7} & 1.83  \\ 
            \footnotesize{VQDIF-only}           & 0.79 & \textbf{73.8} & 0.25 & 3.07 & 60.3 & 3.14  \\ 
            \bottomrule
		\end{tabular}
		\end{adjustbox}
	\caption{
	    Quantitative results on ShapeNet with different scan ambiguity.
	    Ours: top-p=0.4 sampling, $\text{Ours}^*$: top-p=0 sampling.
	}
	\label{tab:cmp_implicit}
\end{table}%

%% file: tab/table_partnet.tex
\begin{table}[t!]%

	\begin{minipage}{\columnwidth}
		\begin{adjustbox}{width=\columnwidth,center}
    		\begin{tabular}{lrrrr}
		        \toprule
        	    \textbf{Method}             & MMD $\downarrow$  & TMD $\uparrow$  & UHD  $\downarrow$ & FPD $\downarrow$ \\ 
		        \midrule
        		 cGAN~\cite{wu2020cGAN}               & 1.98          & 3.05           & 3.39          & 2.95               \\
        		 SInv.~\cite{zhang2021shapeinversion} & 2.14          & 0.62           & 2.32          & 3.45  \\
        		 \textbf{$\text{Ours}$}               & \textbf{1.32} & \textbf{3.96}  & \textbf{0.98} & \textbf{1.22} \\ 
		        \bottomrule
    		\end{tabular}
		\end{adjustbox}
	\end{minipage}
\caption{
	 Quantitative comparison for multi-modal completion on PartNet between our method and prior works.
	 The metrics are averaged across all three categories (Table, Chair, Lamp) and are scaled by $10^{3}$, $10^{2}$, $10^{2}$, $10^{1}$ respectively.
	 }
\label{tab:cmp_partnet}
\end{table}%


%% file: fig/results_realscan.tex
\begin{figure}[t!]
    \centering
    \includegraphics[width=\linewidth]{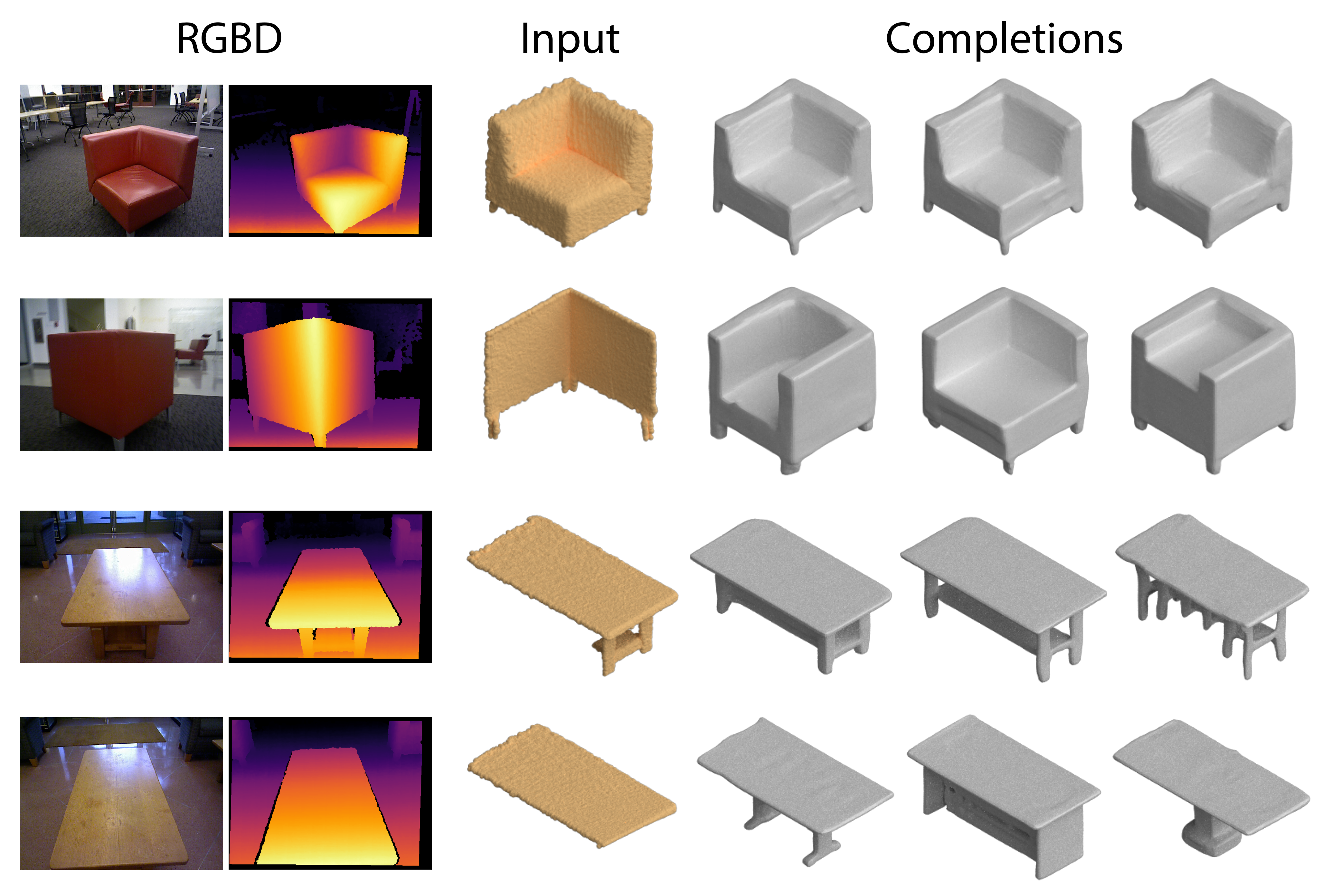}
    \caption{Shape completion results on real-world depth scan from Redwood dataset. ShapeFormer takes partial point clouds converted from depth images and produces multiple possible completions whose variation depends on the uncertainty of viewpoints.
    }
    \label{fig:results_realscan} 
\end{figure}

%% file: fig/results_famous.tex
\begin{figure}[t]
    \centering
    \includegraphics[width=\linewidth]{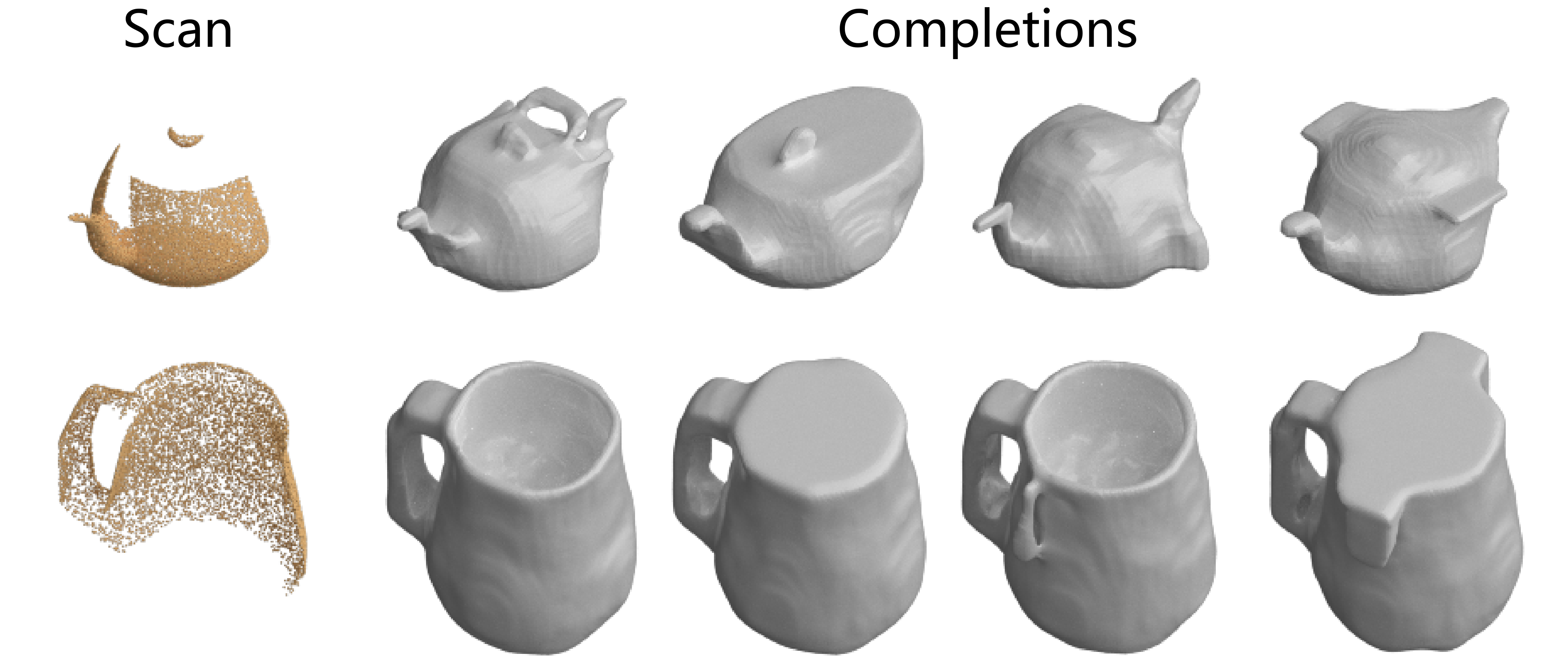}
    \caption{Shape completion results on out-of-distribution shapes. 
    Given a scan of an unseen type of shape, ShapeFormer can produce multiple reasonable completions by generalizing the knowledge learned in the training set.
    }
    \label{fig:results_famous} 
\end{figure}


%% file: fig/results_human.tex
\begin{figure}[t!]
    \centering
    \includegraphics[width=\linewidth]{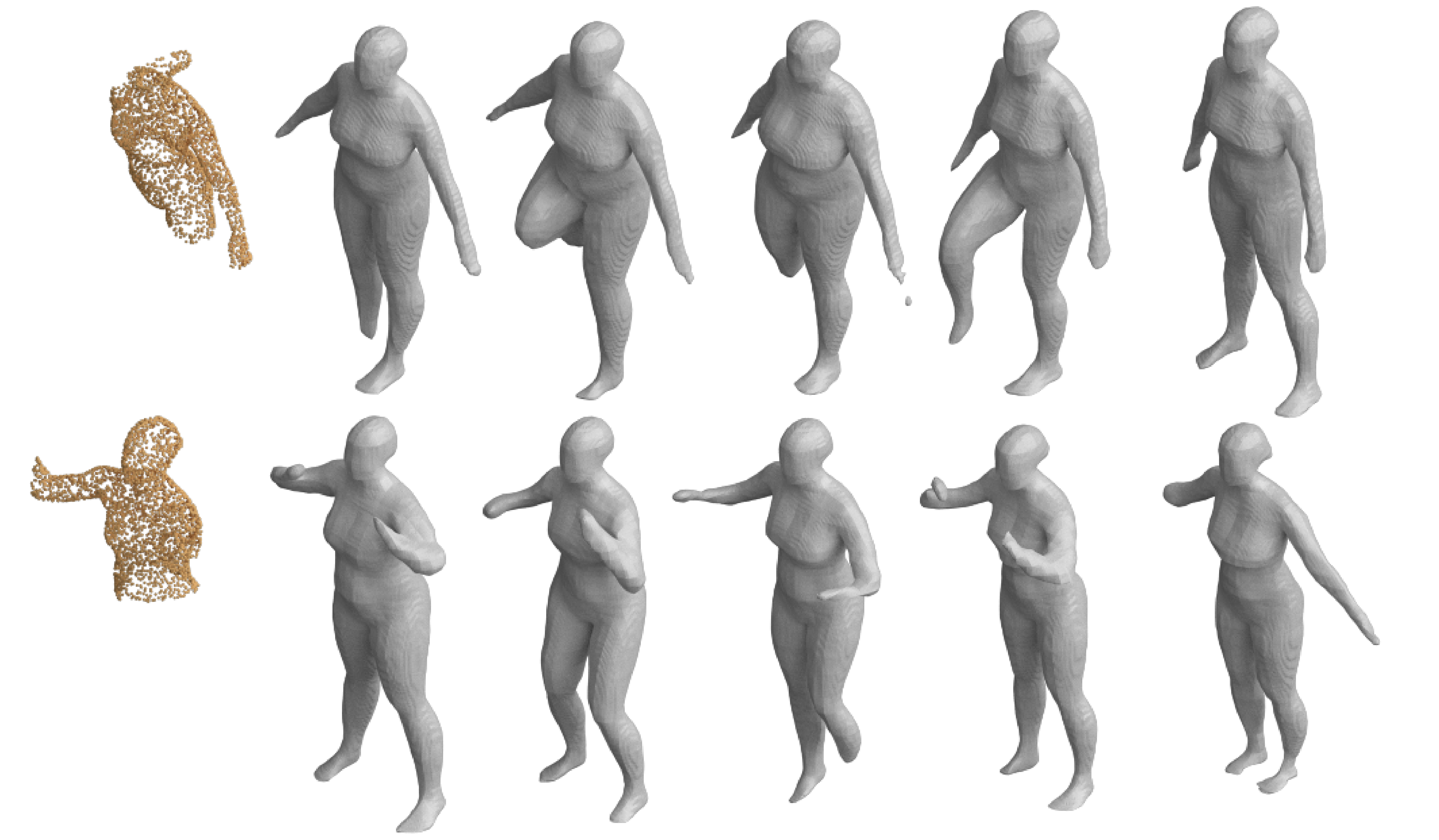}
    \caption{
    Given partial human body parts (left column), our method generates complete human bodies with different poses (along the rows) and the variety depends on the ambiguity.
    }
    \label{fig:results_human} 
\end{figure}

%% file: tab/table_cmp_recon.tex
\begin{table}[t!]%
		\begin{adjustbox}{width=\columnwidth,center}
    		\begin{tabular}{lcccccccc}
        		\toprule 
        	    \textbf{}             & Occ. & CONet. & IF. & $\text{Ours}_8$ & $\text{Ours}_{16}$  & $\text{Ours}_{32}$  \\ 
        	    \midrule
        		 CD       & 3.56 & 0.98 & 0.43 & 1.90 & 0.98 & 0.55 \\
        		 F1       & 68.2 & 89.0 & 97.8 & 77.5 & 88.1 & 96.4 \\
        		 len.     &  1   &  $32^3$   &  $128^3$   & 57   & 217  &  889  \\
        		 \bottomrule
    		\end{tabular}
		\end{adjustbox}
	\caption{
	    Auto-encoding results for objects in ShapeNet. 
	    len. stands for sequence length of the flattened representation.
	 }
	\label{tab:cmp_recon}
\end{table}%

%% file: fig/results_recon_bytes.tex
\begin{figure}[!t]
    \centering
    \includegraphics[width=\linewidth]{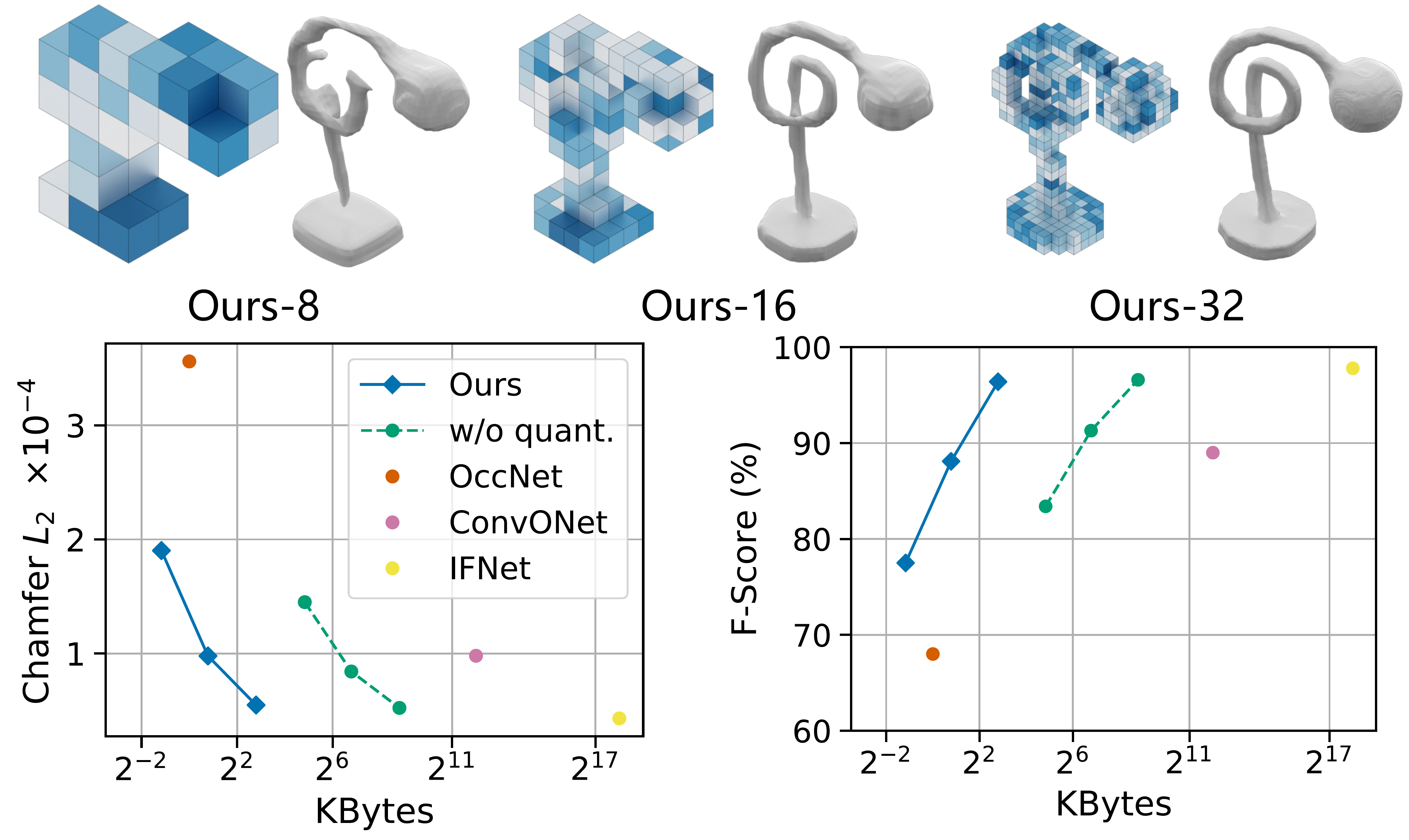}
    \caption{The relation between representation size and reconstruction accuracy. With higher feature resolution, our VQDIF achieves satisfactory accuracy while keeping a rather small byte size.
    }
    \label{fig:results_recon_bytes} 
\end{figure}

%% file: fig/results_recon_cmp.tex
\begin{figure}[t!]
    \centering
    \includegraphics[width=\linewidth]{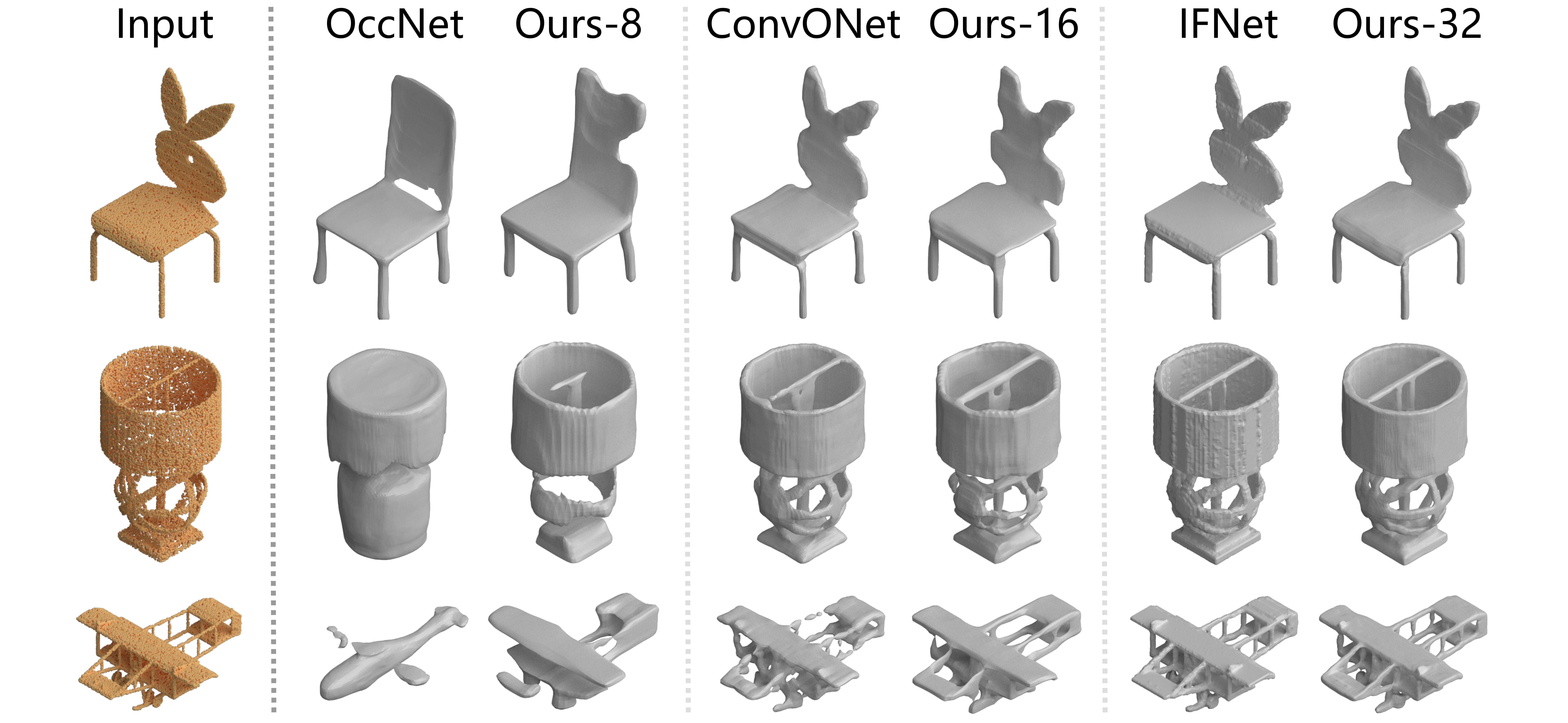}
    \caption{Results for auto-encoding complete shapes. Our VQDIF in different feature resolutions achieves better or similar results compared to the prior DIF methods.
    }
    \label{fig:results_recon_cmp} 
\end{figure}

%% file: sec/5_conclusions.tex
\section{Conclusions}
We have presented ShapeFormer, a transformer-based architecture that learns a conditional distribution of completions, from which multiple plausible completed shapes can be sampled. By explicitly modeling the underlying distribution, our method produces sharp output instead of regressing to the mean producing a blurry result. 
To facilitate generative learning for 3D shape, we propose a new 3D representation VQDIF that can significantly compress the shapes into short sequences of sparse, discrete local features, which in turn enables producing better results, both in terms of quality and diversity, than previous methods.

The major factor limiting our method to be applied in fields like robotics is the sampling speed, which is currently 20 seconds per generated complete shape. In the future, we would also like to explore utilizing a more efficient attention mechanism to allow Transformers to learn VQDIF with smaller size, producing even higher quality completions. Moreover, the current method is generic, leveraging advances in language models. More research is required to include geometric or physical reasoning in the process to better deal with ambiguities.

%% file: sec/6_acknowledgements.tex
{\small
\paragraph{Acknowledgements}
We thank the reviews for their comments. We thank Ziyu Wan, Xuelin Chen and Jiahui Lyu for discussions.
This work was supported in parts by NSFC (62161146005, U21B2023, U2001206), GD Talent Program (2019JC05X328), GD Science and Technology Program (2020A0505100064), DEGP Key Project (2018KZDXM058, 2020SFKC059), Shenzhen Science and Technology Program (RCJC20200714114435012, JCYJ20210324120213036), Royal Society
(NAF-R1-180099), ISF (3441/21, 2492/20) and Guangdong Laboratory of Artificial Intelligence and Digital Economy (SZ).
}

%% file: sec/X_supplementary.tex
\appendix

\setcounter{page}{1}


\twocolumn[
\centering
\Large
\textbf{ShapeFormer: Transformer-based Shape Completion via Sparse Representation} \\
\vspace{0.5em}Supplementary Material \\
\vspace{1.0em}
\small
Xingguang Yan$^{1}$\hspace{.01cm}
Liqiang Lin$^{1}$\hspace{.01cm}
Niloy J. Mitra$^{2,3}$\hspace{.01cm}
Dani Lischinski$^{4}$\hspace{.01cm}
Daniel Cohen-Or$^{1,5}$\hspace{.01cm}
Hui Huang$^{1*}$\\
$^{1}$Shenzhen University\hspace{.02cm}
$^{2}$University College London\hspace{.02cm}
$^{3}$Adobe Research\hspace{.02cm}
$^{4}$Hebrew University of Jerusalem\hspace{.02cm}
$^{5}$Tel Aviv University\\
\vspace{1.0em}
] 
\appendix

\input{fig/supp_ambiguity}
\input{fig/supp_pipeline_vqdif}
\begin{abstract}
In this supplementary document, we first give a detailed description of the ambiguity measure, model architectures, and training/testing statistics in \cref{sec:details}. Then we show more visual comparisons between our method and previous methods for scans of both \textbf{high and low ambiguity} in \cref{sec:more_cmp}. Lastly, we will give more analysis on our method in \cref{sec:more_analysis}, such as a discussion of limitations. The code of our model is also included in the supplementary material.
\end{abstract}

\section{Implementation Details}\label{sec:details}
\subsection{Ambiguity measure for partial point cloud} 
The ambiguity for a partial point cloud measures the variety of its potential complete shapes.
However, the direct measurement for ambiguity is difficult, if not impossible.
In contrast, the incompleteness of a partial cloud toward its complete shape is relatively easy to compute. Although it can not fully reflect ambiguity (e.g., a top scan of a table as incomplete as a bottom scan could have a much greater ambiguity), the ambiguity is still strongly correlated.
\setlength{\columnsep}{0.3cm}%
\begin{wrapfigure}{r}{0.4\linewidth}
    \includegraphics[width=1.0\linewidth]{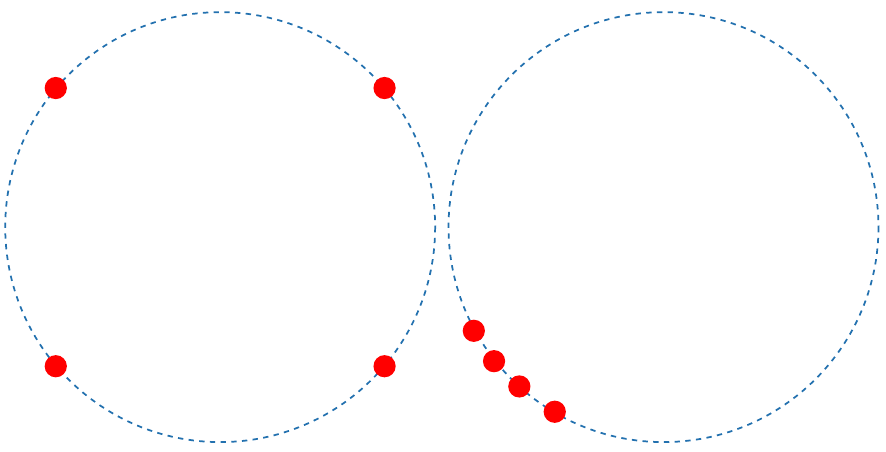}
\end{wrapfigure}
Hence, we seek to find a metric on the incompleteness of such a point cloud to indicate its ambiguity.
Intuitively, we could use metrics like F-score\cite{tatarchenko2019single} to measure the ratio of the approximate partial surface area toward the complete area. But as indicated in the inset figure, such measures will fail to differentiate the coverage difference of the partial cloud (red dots) to the complete one (in blue).
Instead, we propose to use a metric based on Chamfer-$L_2$, which goes larger as the partial point cloud misses more global structure. Since the partial to complete distance is always negligible, we can only calculate the complete to partial distance.
And to compare the ambiguity of scans on different shapes, we normalize the distance of a point according to its farthest distance in the complete shape.
More specifically, we define the metric $Amb$ evaluating the ambiguity of scan $\mathcal{C}$ given the complete point cloud as $\mathcal{B}$ as: 
\begin{equation}
    Amb(B,C) = \frac{1}{B}\Sigma_{x\in \mathcal{B}} \frac{\min_{y\in \mathcal{C}}||x-y||}{\max_{x'\in B}||x-x'||},
\end{equation}
Where $B$ is the number of points in the complete cloud.

We sample 70 views for each shape, 64 of which are evenly sampled from the view sphere (via Fibonacci sampling), and the rest are the six orthogonal views. Then we sort these views according to the score.
In \cref{fig:supp_ambiguity}, we use a teapot as an example to show the score distribution of these 70 scans.
For scans with low ambiguity scores, the underlying shape's global structure is either captured or is clearly indicated by the captured shape salient features.
For example, the scan covering the teapot's mouth, handle, and body can be completed easily.
However, it would be more difficult to infer the complete shape when the score is high since it may have different global structures, and a single explanation is not satisfactory.
As shown in the main paper, our method can better handle such scans than existing shape completion methods.

\input{fig/supp_pipeline_shapeformer}

\input{tab/supp_table_architecture}

\subsection{Architectures} \label{sec:architecture}
We show the detailed architecture of VQDIF and ShapeFormer in \cref{fig:supp_pipeline_vqdif,fig:supp_pipeline_shapeformer}, respectively. and the parameters of their sub-modules are listed in \cref{tab:architecture_details}

\paragraph{VQDIF}
As shown in Figure~\ref{fig:supp_pipeline_vqdif}, VQDIF is an encoder-decoder architecture, where the encoder maps an input point cloud to a discrete sequence representation $\mathcal{S}$, while the decoder maps such a sequence to a deep implicit function $f(\mathbf{x})$.
Unlike the main paper's completion pipeline, both the encoder and decoder only take complete input during training.
The input to the encoder is a point cloud $\mathcal{P}\in\mathbb{R}^{N\times3}$ representing the dense sampling of a shape or its partial observation. 
During the training phase, we use complete dense clouds with $N=32768$ points to train VQDIF to capture local geometric details in the input.
At test time, we use the trained encoder to directly encode partial point clouds, which may be sparse or dense.

The encoder first processes the input cloud with a local pooled PointNet~\cite{qi2017pointnet} to obtain a feature grid.
Similar to prior work \cite{Peng2020ConvONet}, the local pooled PointNet aggregates features within a grid cell in contrast to the original PointNet, where all point features are pooled together to obtain a global feature. Specifically, we use a grid of resolution $64$ with a feature size of $32$.

Next, to reduce the number of local features, the high-resolution feature grid is down-sampled to lower resolution $R$, using several consecutive strided convolution blocks.
As shown in \cref{tab:architecture_details}, the parameters of these blocks are carefully set to have the least receptive field since a large receptive field lets each grid feature cover a larger region, reducing the sparsity of the representation.
We can then extract the non-empty features by directly masking the encoded feature grid with the voxelized input point cloud (resolution $R$) thanks to the minimum receptive field.
After flattening and quantizing the features (see the main paper), we get the 2-tuple sequence representation directly sent to the decoder. Note that we also save the "empty" feature to project the sequence back to the feature grid in the decoder.

The decoder consists of a 3D U-Net \cite{cciccek2016unet3d}, an up-sampler, and an implicit decoder. It first projects the quantized sparse sequence back to a 3D feature grid, which serves as the input for the 3D U-Net.
In contrast to the encoder, the decoder is designed to have a large receptive field. 
This is because, in order for the implicit decoder to infer whether a probe lies inside or outside of the shape, we need global knowledge. This is in alignment with prior works \cite{Peng2020ConvONet, Erler2020Points2Surf}.
More specifically, we use a 3-step U-Net to increase the receptive field, which integrates both local and global information.
The up-sampler has the same number of scaling stages as the down-sampler, but it has a larger receptive field by design.
Lastly, similarly to prior work \cite{Peng2020ConvONet}, the implicit decoder consists of multiple ResNet blocks.
It takes querying probe points $\mathcal{T}_\mathbf{x}$ and predicts their occupancy probability $\mathcal{T}_{o}$.

\paragraph{ShapeFormer}
In \cref{fig:supp_pipeline_shapeformer}, we show the detailed architecture of ShapeFormer. 
The input to the ShapeFormer consists of the concatenated sequence of $\mathcal{S}_\mathcal{P}$ and $\mathcal{S}_\mathcal{C}$.
Since these sequences both have variable lengths, we append an end-token ([END]) to each sequence to indicate when the sequence terminates. 
Next, as in prior works \cite{nash2021DCTransformer, dieleman2021SlowAEs}, all these indices are turned into learnable embeddings and are additively combined as the input embedding for ShapeFormer.

The main components of ShapeFormer are two causally-masked transformers, which consist of multiple decoder-only transformer blocks \cite{radford2019GPT2}. 
The first transformer learns to predict the coordinate of the next tuple, conditioned on previous tuples, while
the second one learns to predict the value of the next element conditioned on previous tuples and the (predicted) coordinate index of the next element. Thus, the output feature of the first transformer is additively mixed with the input embedding of the second transformer delivering the encoded sequence information.

Each transformer is followed by an output head, which converts the feature produced by the transformer into a categorical distribution of the next sequence element.
Both output heads consist of two fully connected layers, followed by a softmax layer to produce categorical conditional distributions for each of the sequence elements: $\{(p_{c_i}, p_{v_i})\}_{i=1}^K$. Note that this essentially shifts the complete sequence to the right by one element. 
For training, we also empirically find randomly masking out the partial sequence will improve generalization.

\subsection{Details on training and sampling} \label{sec:training}
We use Adam optimizer for training both VQDIF and ShapeFormer, and we set the learning rate as $1e-4$ for VQDIF and $1e-5$ for ShapeFormer.
We use step decay for VQDIF with step size equal to 10 and $\beta=.9$ and do not apply learning rate scheduling for ShapeFormer.
We train our network on a deep learning server with Intel Xeon CPU E5-2680 v4 CPU*56 and 256GB memory with 10 Nvidia Quadro P6000 graphics cards with a GPU memory size of 24GB.
It takes 30 hours for our model to converge on our virtual scan dataset and 8 hours on the PartNet dataset.
For D-Faust, the converging time is 16 hours.
For sampling, we can obtain a single sample sequence in roughly 20 seconds, and we can also sample 24 sequences in parallel in 5 minutes.

\input{fig/supp_more_comparisons}
\input{fig/supp_more_high}
\input{fig/supp_more_low}

\section{More comparisons} \label{sec:more_cmp}
We show more visual comparisons between our method and prior state-of-the-art methods in \cref{fig:supp_more_comparisons,fig:supp_more_high,fig:supp_more_low}.
\cref{fig:supp_more_comparisons,fig:supp_more_high} illustrates results on high-ambiguity scans, 
In these examples, we can see the averaging effect of the deterministic methods (See the scattering effect in ambiguous regions of the completions of PoinTr~\cite{yu2021pointr}).
Our method produces significantly better results in terms of quality and diversity.

Also, we demonstrate our method can also achieve competitive accuracy for low-ambiguity scans in \cref{fig:supp_more_low}.
Since there is limited ambiguity for such scans and the goal is to achieve accuracy toward ground truth, 
we put the ground truth in the first row and only sample 1 completion for each of our sampling strategies (Ours: top-.4 sampling, Ours*: top-.0, e.g., best sampling).
Also, we only compare state-of-the-art deterministic methods: ConvONet~\cite{Peng2020ConvONet}, IF-Net~\cite{chibane20ifnet}, and PoinTr~\cite{yu2021pointr} in these examples.
As we can see, even the scans cover most areas of the ground truth shape; prior works can still produce unsatisfactory results for unseen regions.
In contrast, our method can always produce more accurate, high-quality completions.
Moreover, since Ours* always picks the coordinate and value indices with the highest probability, it often produces slightly more accurate shapes.

\section{More analysis} \label{sec:more_analysis}
\paragraph{Discussion of Limitation}
ShapeFormer inherits the typical limitations of transformer-based autoregressive models. Mainly, the representation length cannot be too long, and thus the method currently can only use VQDIF with $R=16$, which may fail to complete and reconstruct shapes with intricate structures; an example is shown in Figure~\ref{fig:supp_failure}. 
Another related limitation is the sampling speed, which prevents interactive applications. 
\input{fig/supp_failure}

There are two research avenues to alleviate these problems:
(i)~Investigating more efficient attention mechanisms to reduce the transformer's quadratic complexity in the sequence length $K$ to $O(K\sqrt{K})$ \cite{ho2019axial} or even $O(K)$ \cite{choromanski2021performer}. 
(ii)~Designing an adaptive quantization scheme for the point clouds, which enables Transformers to focus dependencies on a lower local level while using higher-level features for faraway regions.
(iii) Adopt advanced sampling techniques for autoregressive models such as parallel sampling~\cite{jayaram2021parallel}.

Moreover, since we generate sequences of complete shapes from scratch, our results may slightly alter the input geometry to overcome the potential sparsity and noise. Besides using higher resolution quantized features to obtain more accurate generation, another possible improvement to this issue is to include high-resolution features of the input in the decoding procedure as in a recent image inpainting technique~\cite{wan2021ict}.

%% file: fig/supp_ambiguity.tex
\begin{figure}[!b]
    \centering
    \includegraphics[width=\linewidth]{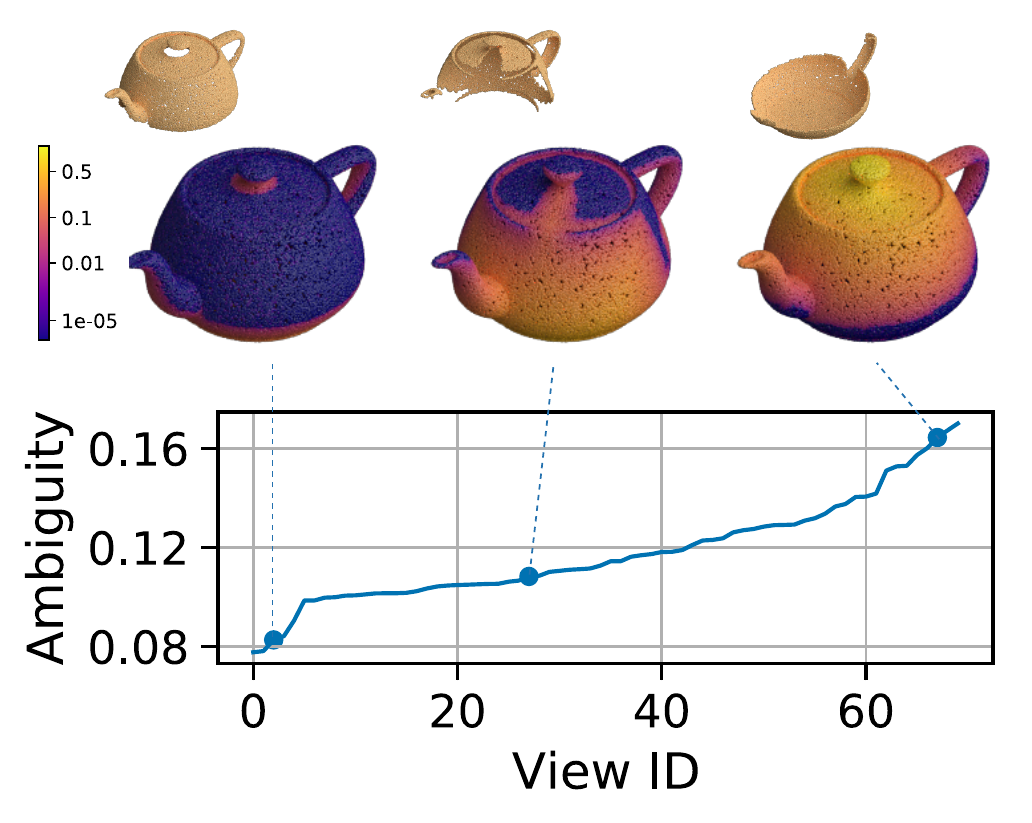}
    \caption{
    Viewing direction greatly influences the scan ambiguity. Our proposed scores for 70 scans of a teapot are shown in sorted order, with examples marked with their position on the curve. The example contains the scans (in gold insets) and complete shape color-coded scores for each point in it.
    }
    \label{fig:supp_ambiguity} 
\end{figure}

%% file: fig/supp_pipeline_vqdif.tex
\begin{figure*}[!t]
    \centering
    \includegraphics[width=\linewidth]{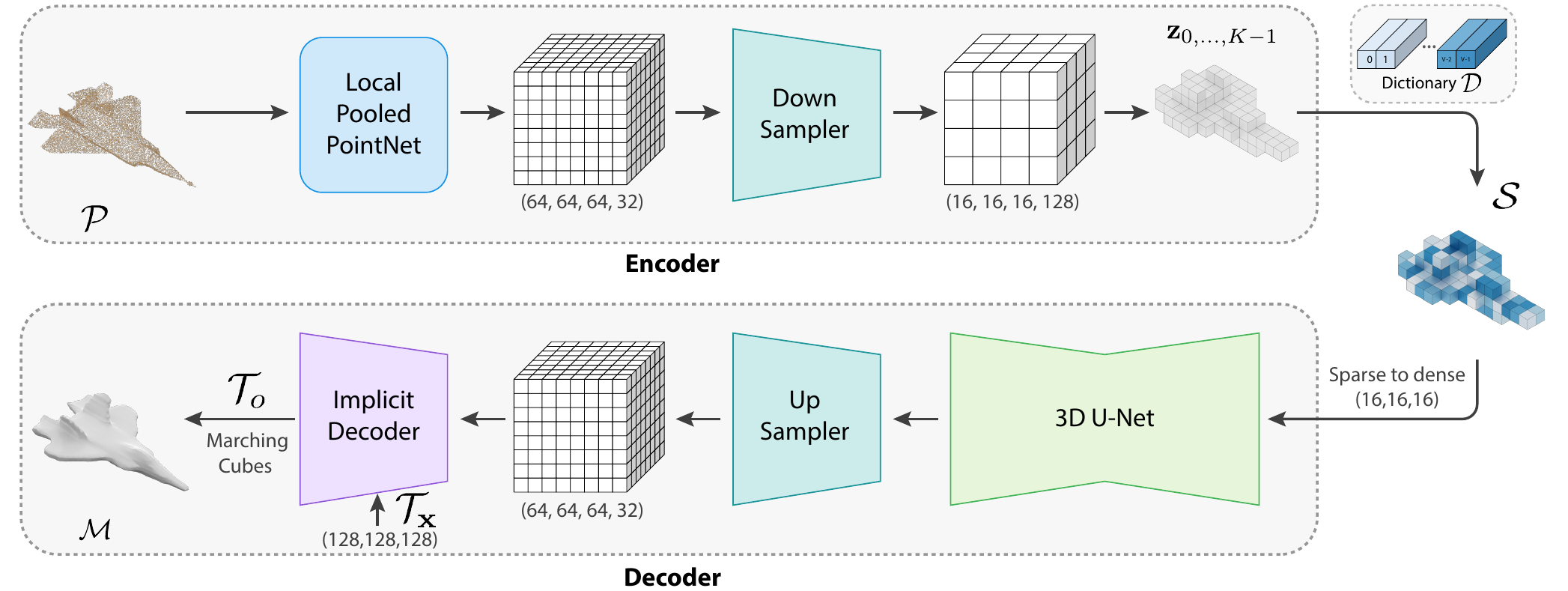}
\caption{
The architecture of VQDIF. The complete point cloud $\mathcal{P}$ is encoded to a feature grid and down-sampled into a lower resolution one. Its non-empty features are then flattened and quantized to form the VQDIF sequence which is then projected back to a feature grid, up-sampled and sent to an implicit decoder, from which the occupancy grid $\mathcal{T}_{o}$ of probes $\mathcal{T}_\mathbf{x}$ and the reconstruction $\mathcal{M}$ can be obtained.
}
\label{fig:supp_pipeline_vqdif} 
\end{figure*}


%% file: fig/supp_pipeline_shapeformer.tex
\begin{figure}[!t]
    \centering
    \includegraphics[width=\linewidth]{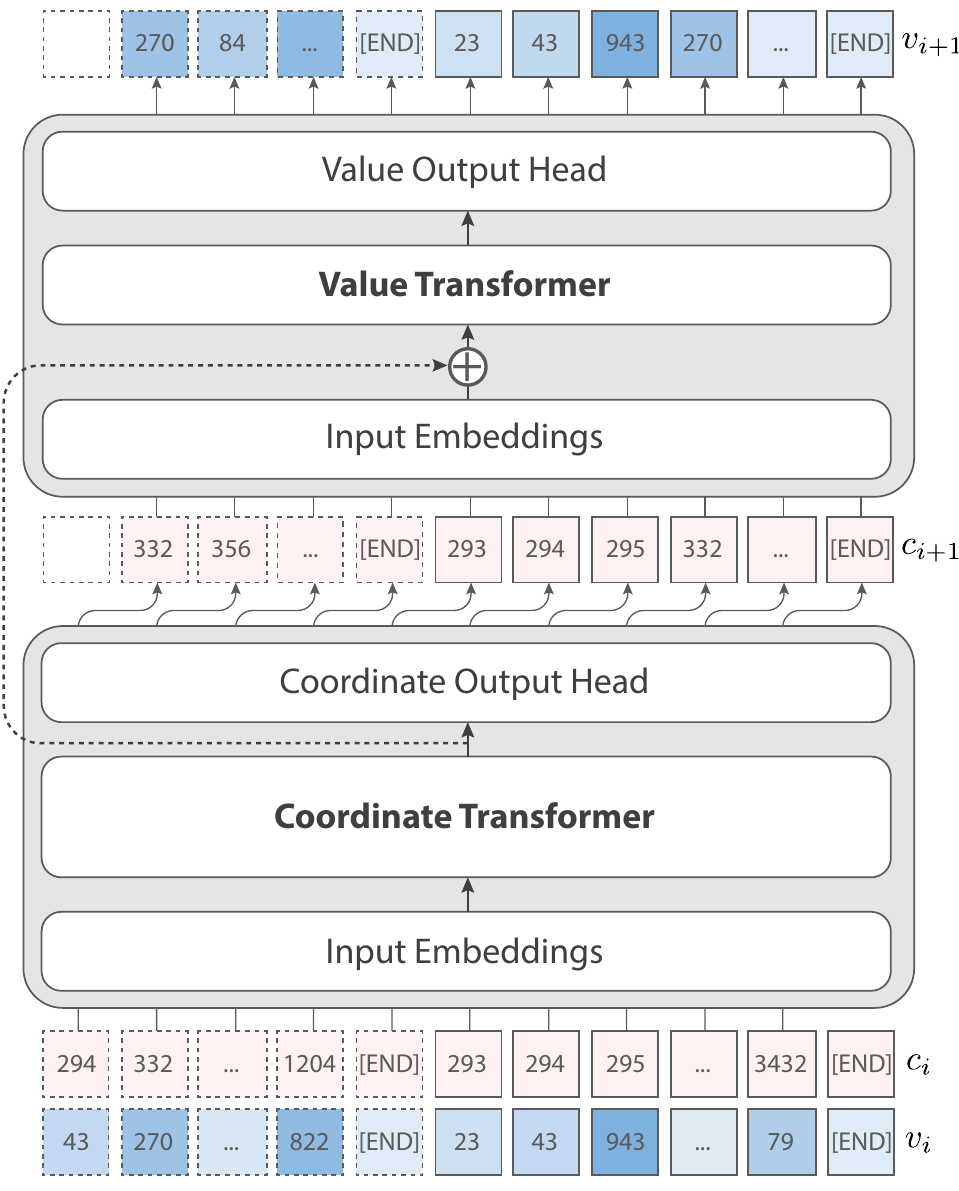}
\caption{
An extended view of ShapeFormer. Different from the figure in the main paper, we show the inside of each Transformer module.
The input embeddings are obtained by additively mixing the location and value embeddings.
And the output head converts the output embedding into categorical distributions.
}

\label{fig:supp_pipeline_shapeformer} 
\end{figure}

%% file: tab/supp_table_architecture.tex
\begin{table}[t!]

    \begin{adjustbox}{width=\columnwidth,center}	
		\begin{tabular}{lll} \hline
			Layer Name   & Notes & Input Size \\ \hline\hline

			\textbf{VQDIF } 					   &  &\\
    			Local Pooled Pointnet &      & $N\times 3$\\
    			Downsampler &         &\\
        			\quad ConvLayer & k2s2p0     & $64\times64\times64\times32$\\
        			\quad ConvLayer & k1s1p0     & $32\times32\times32\times64$\\
        			\quad ConvLayer & k2s2p0     & $64\times64\times64\times32$\\
        			\quad ConvLayer & k1s1p0     & $32\times32\times32\times64$\\
    			Quantizer                        &  & $16\times16\times16\times128$\\  
    			UNet3D &           & $16\times16\times16\times128$ \\
    			Upsampler &        & $16\times16\times16\times128$ \\
        			\quad Scaling   & nearest mode     & $16\times16\times16\times128$\\
        			\quad ConvLayer & k3s1p1     & $32\times32\times32\times128$\\
        			\quad ConvLayer & k3s1p1     & $32\times32\times32\times64$\\
        			\quad Scaling   & nearest mode     & $32\times32\times32\times64$\\
        			\quad ConvLayer & k3s1p1     & $64\times64\times64\times64$\\
        			\quad ConvLayer & k3s1p1     & $64\times64\times64\times32$\\
    			Upsampler Output&        & $64\times64\times64\times32$ \\
    			Implicit Decoder   & & $128^3\times3$\\
    			Implicit Decoder Output       & & $128^3\times1$\\
			\hline\hline
			\textbf{ShapeFormer}				&   & \\
    		Embedding Blocks    &   \#4M      & $K\times 2$\\
    		Coordinate Transformer Blocks $\times 20$     & \#251M & $K\times 1024$\\
    		Coordinate Output Heads &           \#4M  & $K\times 4097$\\
    		Embedding Blocks               &    \#4M      & $K\times 2$\\
    		Value Transformer Blocks $\times 4$ & \#50M  & $K\times 1024$\\
    		Value Output Heads &                \#4M  & $K\times 4097$\\

			\hline\hline
			\textbf{Total params}     & \#340M &\\
			\textbf{Trainable params} & \#323M &\\
			\hline
		\end{tabular}
    \end{adjustbox}
	\caption{The detailed architecture information of our method. $N$ is the point size.
	For both VQDIF and ShapeFormer, we list the input size of their components.
	For convolutional neural networks, the "k", "s", "p" stands for kernel size, stride, and padding, respectively.
	Also "ConvLayer" denotes the composition of CNN + ReLU + GroupNorm.
	We also list the number of parameters for each component and indicate them with \#.
	The sequence length is denoted by K, with a maximum of $812$.
	}
	\label{tab:architecture_details}
\end{table}

%% file: fig/supp_more_comparisons.tex
\begin{figure*}[!t]
    \centering
    \includegraphics[width=\textwidth]{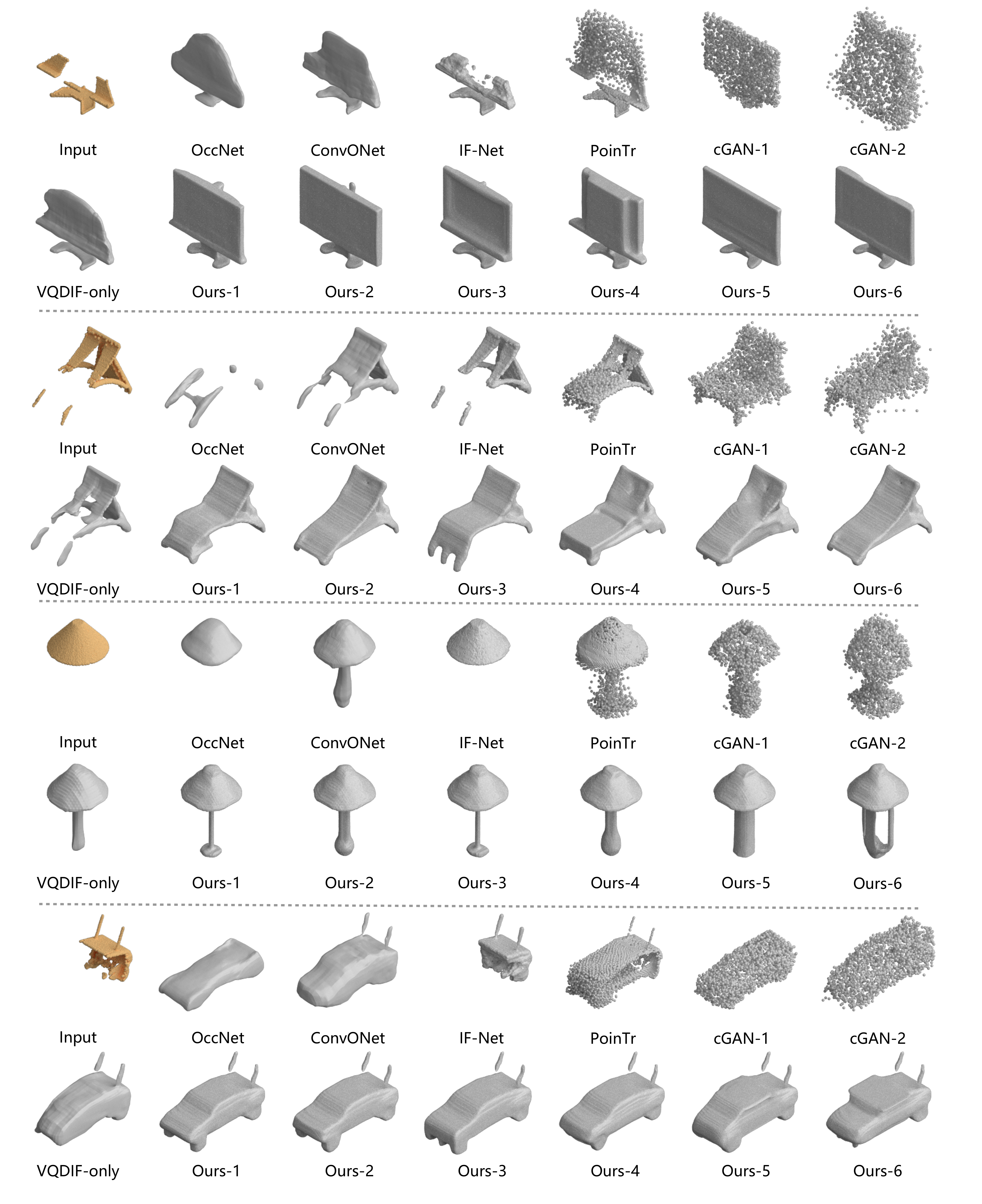}
    \caption{
    More comparisons on high ambiguity scans of ShapeNet objects.
}
    \label{fig:supp_more_comparisons} 
\end{figure*}

    

%% file: fig/supp_more_high.tex
\begin{figure*}[!t]
    \centering
    \includegraphics[width=\textwidth]{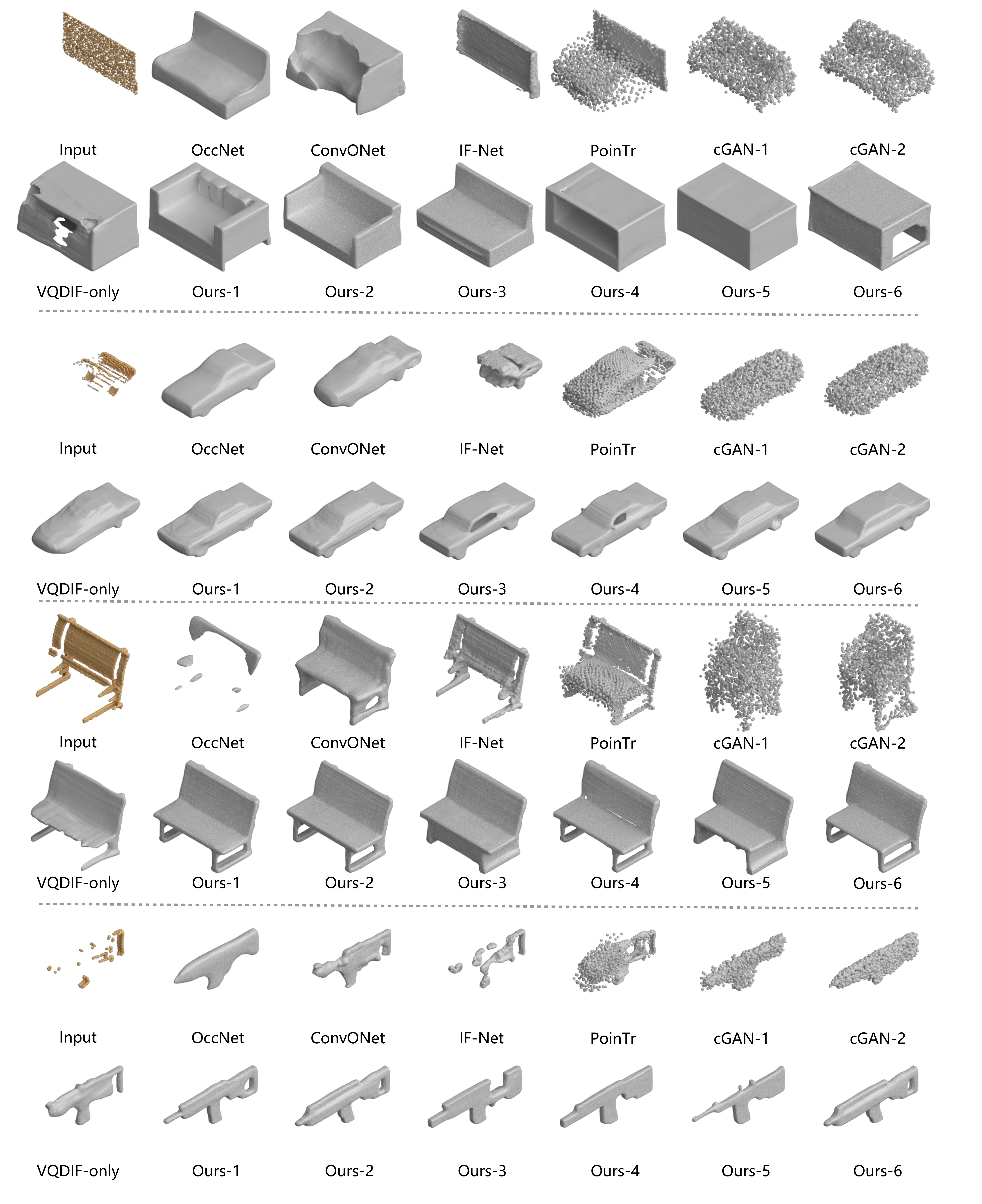}
    \caption{
    More comparisons on high ambiguity scans of ShapeNet objects.
}\label{fig:supp_more_high} 
\end{figure*}


%% file: fig/supp_more_low.tex
\begin{figure*}[!t]
    \centering
    \includegraphics[width=\textwidth]{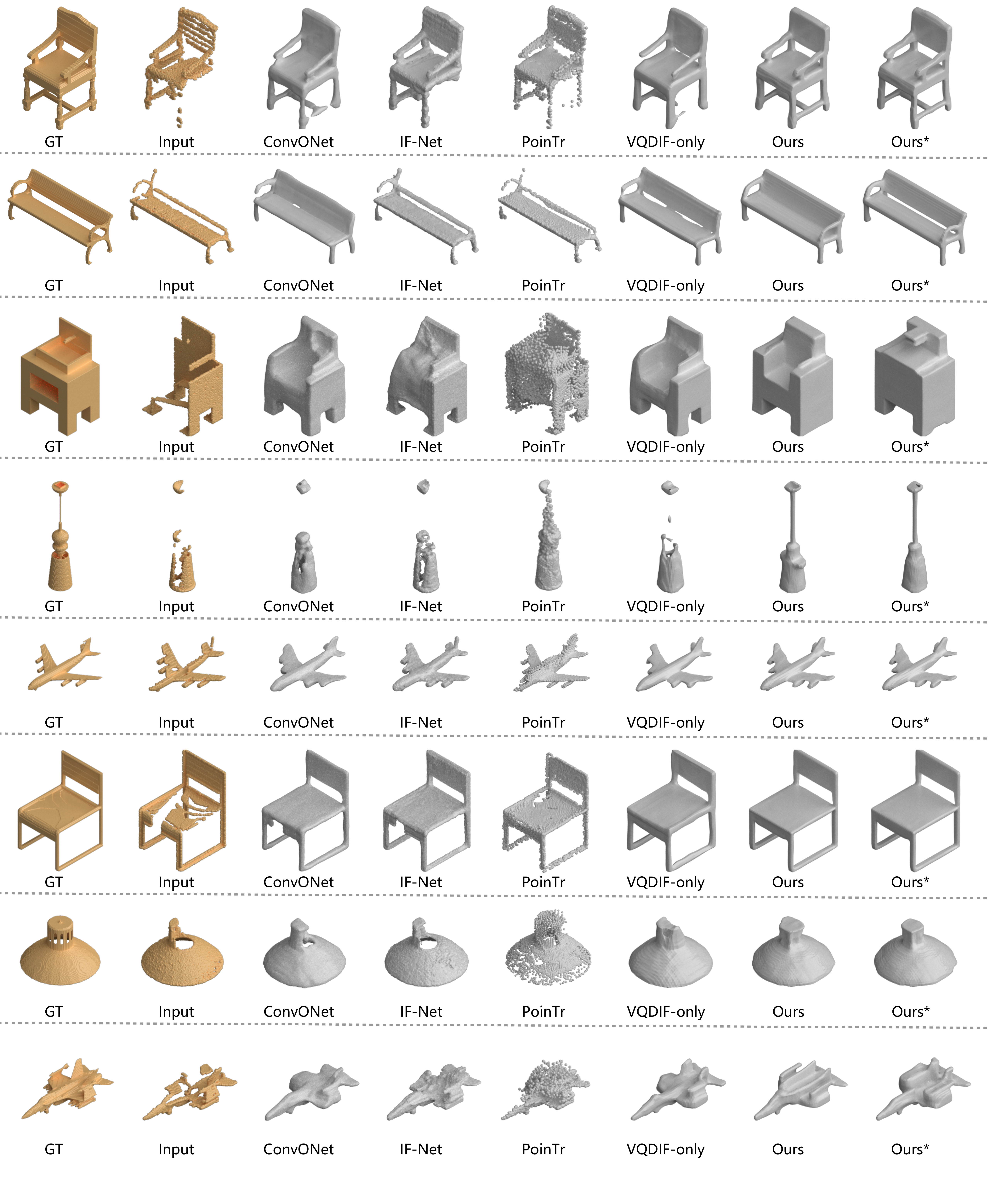}
    \caption{
    More comparisons on low ambiguity scans of ShapeNet objects. Ours=top-.4 sampling, Ours*=top-.0 sampling (best sampling).
} \label{fig:supp_more_low} 
\end{figure*}


%% file: fig/supp_failure.tex
\begin{figure}[h]
    \centering
    \includegraphics[width=\linewidth]{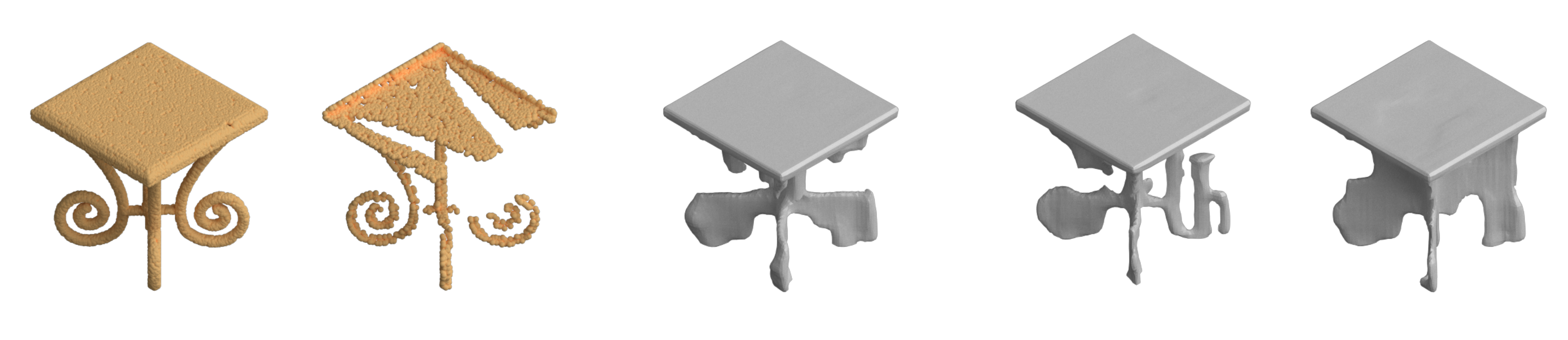}
    \caption{An example of a shape completion failure case of ShapeFormer.
    The intricate details present in the input (second from left) are not preserved in the completions (gray shapes). The leftmost image shows the ground truth shape. }
    \label{fig:supp_failure} 
\end{figure}

%% file: main.bbl
\begin{thebibliography}{10}\itemsep=-1pt

\bibitem{Achlioptas2017pointae}
Panos Achlioptas, Olga Diamanti, Ioannis Mitliagkas, and Leonidas~J. Guibas.
\newblock Learning representations and generative models for {3D} point clouds.
\newblock In {\em International conference on machine learning}, 2017.

\bibitem{arora2021multimodal}
Himanshu Arora, Saurabh Mishra, Shichong Peng, Ke Li, and Ali Mahdavi-Amiri.
\newblock Multimodal shape completion via imle.
\newblock {\em arXiv preprint arXiv:2106.16237}, 2021.

\bibitem{bao2021beit}
Hangbo Bao, Li Dong, and Furu Wei.
\newblock Beit: Bert pre-training of image transformers, 2021.

\bibitem{bengio2000modeling}
Yoshua Bengio and Samy Bengio.
\newblock Modeling high-dimensional discrete data with multi-layer neural
  networks.
\newblock {\em Advances in Neural Information Processing Systems}, 12:400--406,
  2000.

\bibitem{berger2017survey}
Matthew Berger, Andrea Tagliasacchi, Lee~M Seversky, Pierre Alliez, Gael
  Guennebaud, Joshua~A Levine, Andrei Sharf, and Claudio~T Silva.
\newblock A survey of surface reconstruction from point clouds.
\newblock In {\em Computer Graphics Forum}, volume~36, pages 301--329, 2017.

\bibitem{bernardini1999ball}
Fausto Bernardini, Joshua Mittleman, Holly Rushmeier, Claudio Silva, and
  Gabriel Taubin.
\newblock The ball-pivoting algorithm for surface reconstruction.
\newblock {\em IEEE transactions on visualization and computer graphics},
  5(4):349--359, 1999.

\bibitem{federica2017CVPRdfaust}
Federica Bogo, Javier Romero, Gerard Pons-Moll, and Michael~J. Black.
\newblock Dynamic {FAUST}: {R}egistering human bodies in motion.
\newblock In {\em IEEE Conf. on Computer Vision and Pattern Recognition
  (CVPR)}, July 2017.

\bibitem{brown2020GPT3}
Tom~B Brown, Benjamin Mann, Nick Ryder, Melanie Subbiah, Jared Kaplan, Prafulla
  Dhariwal, Arvind Neelakantan, Pranav Shyam, Girish Sastry, Amanda Askell,
  et~al.
\newblock Language models are few-shot learners.
\newblock {\em arXiv preprint arXiv:2005.14165}, 2020.

\bibitem{chang2015shapenet}
Angel~X. Chang, Thomas Funkhouser, Leonidas Guibas, Pat Hanrahan, Qixing Huang,
  Zimo Li, Silvio Savarese, Manolis Savva, Shuran Song, Hao Su, Jianxiong Xiao,
  Li Yi, and Fisher Yu.
\newblock Shapenet: An information-rich 3d model repository, 2015.

\bibitem{qi2017pointnet}
R.~Qi Charles, Hao Su, Mo Kaichun, and Leonidas~J. Guibas.
\newblock Pointnet: Deep learning on point sets for 3d classification and
  segmentation.
\newblock {\em 2017 IEEE Conference on Computer Vision and Pattern Recognition
  (CVPR)}, Jul 2017.

\bibitem{chen2020imagegpt}
Mark Chen, Alec Radford, Rewon Child, Jeffrey Wu, Heewoo Jun, David Luan, and
  Ilya Sutskever.
\newblock Generative pretraining from pixels.
\newblock In {\em International Conference on Machine Learning}, pages
  1691--1703. PMLR, 2020.

\bibitem{chen2018pixelsnail}
Xi Chen, Nikhil Mishra, Mostafa Rohaninejad, and Pieter Abbeel.
\newblock Pixelsnail: An improved autoregressive generative model.
\newblock In {\em International Conference on Machine Learning}, pages
  864--872. PMLR, 2018.

\bibitem{zhiqin2019imnet}
Zhiqin Chen and Hao Zhang.
\newblock Learning implicit fields for generative shape modeling.
\newblock In {\em Proc. IEEE Conf. on Computer Vision \& Pattern Recognition},
  pages 5939--5948, 2019.

\bibitem{chen2021mdif}
Zhang Chen, Yinda Zhang, Kyle Genova, Sean Fanello, Sofien Bouaziz, Christian
  Hane, Ruofei Du, Cem Keskin, Thomas Funkhouser, and Danhang Tang.
\newblock Multiresolution deep implicit functions for 3d shape representation.
\newblock In {\em Proc. Int. Conf. on Computer Vision}, pages 13087--13096,
  2021.

\bibitem{chibane20ifnet}
Julian Chibane, Thiemo Alldieck, and Gerard Pons-Moll.
\newblock Implicit functions in feature space for 3d shape reconstruction and
  completion.
\newblock In {\em Proc. IEEE Conf. on Computer Vision \& Pattern Recognition}.
  {IEEE}, jun 2020.

\bibitem{choi2016redwood}
Sungjoon Choi, Qian-Yi Zhou, Stephen Miller, and Vladlen Koltun.
\newblock A large dataset of object scans, 2016.

\bibitem{choromanski2021performer}
Krzysztof Choromanski, Valerii Likhosherstov, David Dohan, Xingyou Song,
  Andreea Gane, Tamas Sarlos, Peter Hawkins, Jared Davis, Afroz Mohiuddin,
  Lukasz Kaiser, David Belanger, Lucy Colwell, and Adrian Weller.
\newblock Rethinking attention with performers, 2021.

\bibitem{choy20163dr2n2}
Christopher~B Choy, Danfei Xu, JunYoung Gwak, Kevin Chen, and Silvio Savarese.
\newblock 3d-r2n2: A unified approach for single and multi-view 3d object
  reconstruction.
\newblock In {\em Proceedings of the European Conference on Computer Vision
  ({ECCV})}, 2016.

\bibitem{cciccek2016unet3d}
{\"O}zg{\"u}n {\c{C}}i{\c{c}}ek, Ahmed Abdulkadir, Soeren~S Lienkamp, Thomas
  Brox, and Olaf Ronneberger.
\newblock 3d u-net: learning dense volumetric segmentation from sparse
  annotation.
\newblock In {\em International conference on medical image computing and
  computer-assisted intervention}, pages 424--432. Springer, 2016.

\bibitem{dai20173depn}
Angela Dai, Charles~Ruizhongtai Qi, and Matthias NieBner.
\newblock Shape completion using {3D-Encoder-Predictor CNNs} and shape
  synthesis.
\newblock In {\em Proc. IEEE Conf. on Computer Vision \& Pattern Recognition},
  2017.

\bibitem{devlin2018bert}
Jacob Devlin, Ming-Wei Chang, Kenton Lee, and Kristina Toutanova.
\newblock Bert: Pre-training of deep bidirectional transformers for language
  understanding.
\newblock {\em arXiv preprint arXiv:1810.04805}, 2018.

\bibitem{dieleman2021SlowAEs}
Sander Dieleman, Charlie Nash, Jesse Engel, and Karen Simonyan.
\newblock Variable-rate discrete representation learning.
\newblock {\em arXiv preprint arXiv:2103.06089}, 2021.

\bibitem{Erler2020Points2Surf}
Philipp Erler, Paul Guerrero, Stefan Ohrhallinger, N. Mitra, and M. Wimmer.
\newblock Points2surf learning implicit surfaces from point clouds.
\newblock In {\em Proc. Euro. Conf. on Computer Vision}, 2020.

\bibitem{esser2020taming}
Patrick Esser, Robin Rombach, and Björn Ommer.
\newblock Taming transformers for high-resolution image synthesis, 2020.

\bibitem{fan2017pointgen}
Haoqiang Fan, Hao Su, and Leonidas Guibas.
\newblock A point set generation network for {3D} object reconstruction from a
  single image.
\newblock In {\em Proc. IEEE Conf. on Computer Vision \& Pattern Recognition},
  2017.

\bibitem{Fauw2019HierarchicalAR}
J. Fauw, S. Dieleman, and K. Simonyan.
\newblock Hierarchical autoregressive image models with auxiliary decoders.
\newblock {\em ArXiv}, abs/1903.04933, 2019.

\bibitem{mvstutorial}
Yasutaka Furukawa and Carlos Hern{\'a}ndez.
\newblock {\em Multi-view stereo: A tutorial}, volume~9.
\newblock 2013.

\bibitem{genova2020ldif}
Kyle Genova, F. Cole, Avneesh Sud, Aaron Sarna, and T. Funkhouser.
\newblock Local deep implicit functions for 3d shape.
\newblock {\em Proc. IEEE Conf. on Computer Vision \& Pattern Recognition},
  pages 4856--4865, 2020.

\bibitem{germain2015made}
Mathieu Germain, Karol Gregor, Iain Murray, and Hugo Larochelle.
\newblock Made: Masked autoencoder for distribution estimation.
\newblock In {\em International conference on machine learning}, pages
  881--889. PMLR, 2015.

\bibitem{groueix2018atlasnet}
Thibault Groueix, Matthew Fisher, Vladimir~G. Kim, Bryan Russell, and Mathieu
  Aubry.
\newblock {AtlasNet: A Papier-M\^ach\'e Approach to Learning {3D} Surface
  Generation}.
\newblock In {\em Proc. IEEE Conf. on Computer Vision \& Pattern Recognition},
  2018.

\bibitem{han2019image}
Xian-Feng Han, Hamid Laga, and Mohammed Bennamoun.
\newblock Image-based 3d object reconstruction: State-of-the-art and trends in
  the deep learning era.
\newblock {\em IEEE transactions on pattern analysis and machine intelligence},
  43(5):1578--1604, 2019.

\bibitem{Hne2017hsp}
Christian H{\"a}ne, Shubham Tulsiani, and Jitendra Malik.
\newblock Hierarchical surface prediction for 3d object reconstruction.
\newblock {\em 2017 International Conference on 3D Vision (3DV)}, pages
  412--420, 2017.

\bibitem{he2021mae}
Kaiming He, Xinlei Chen, Saining Xie, Yanghao Li, Piotr Dollár, and Ross
  Girshick.
\newblock Masked autoencoders are scalable vision learners, 2021.

\bibitem{ho2019axial}
Jonathan Ho, Nal Kalchbrenner, Dirk Weissenborn, and Tim Salimans.
\newblock Axial attention in multidimensional transformers, 2019.

\bibitem{holtzman2019nucleus}
Ari Holtzman, Jan Buys, Li Du, Maxwell Forbes, and Yejin Choi.
\newblock The curious case of neural text degeneration.
\newblock In {\em Proc. Int. Conf. on Learning Representations}, 2019.

\bibitem{huang2019VIPSS}
Zhiyang Huang, Nathan Carr, and Tao Ju.
\newblock Variational implicit point set surfaces.
\newblock {\em ACM Transactions on Graphics (TOG)}, 38(4):1--13, 2019.

\bibitem{jayaram2021parallel}
Vivek Jayaram and John Thickstun.
\newblock Parallel and flexible sampling from autoregressive models via
  langevin dynamics, 2021.

\bibitem{jiang2020lig}
Chiyu Jiang, Avneesh Sud, Ameesh Makadia, Jingwei Huang, Matthias Nie{\ss}ner,
  and Thomas Funkhouser.
\newblock Local implicit grid representations for 3d scenes.
\newblock In {\em Proceedings of the IEEE Conference on Computer Vision and
  Pattern Recognition}, 2020.

\bibitem{kazhdan2006poisson}
Michael Kazhdan, Matthew Bolitho, and Hugues Hoppe.
\newblock Poisson surface reconstruction.
\newblock In {\em Proc. Eurographics Symp. on Geometry Processing}, volume~7,
  2006.

\bibitem{kazhdan2013poisson}
Michael Kazhdan and Hugues Hoppe.
\newblock Screened poisson surface reconstruction.
\newblock {\em ACM Trans. on Graphics}, 32:29:1--29:13, 2013.

\bibitem{litany2018deformable}
Or Litany, Alex Bronstein, Michael Bronstein, and Ameesh Makadia.
\newblock Deformable shape completion with graph convolutional autoencoders,
  2018.

\bibitem{liu2018wikigen}
Peter~J. Liu, Mohammad Saleh, Etienne Pot, Ben Goodrich, Ryan Sepassi, Lukasz
  Kaiser, and Noam Shazeer.
\newblock Generating wikipedia by summarizing long sequences, 2018.

\bibitem{liu2021IMLSNet}
Shi-Lin Liu, Hao-Xiang Guo, Hao Pan, Peng-Shuai Wang, Xin Tong, and Yang Liu.
\newblock Deep implicit moving least-squares functions for 3d reconstruction.
\newblock {\em arXiv preprint arXiv:2103.12266}, 2021.

\bibitem{lars2019occnet}
Lars Mescheder, Michael Oechsle, Michael Niemeyer, Sebastian Nowozin, and
  Andreas Geiger.
\newblock Occupancy networks: Learning {3D}reconstruction in function space.
\newblock In {\em Proc. IEEE Conf. on Computer Vision \& Pattern Recognition},
  pages 4460--4470, 2019.

\bibitem{Mittal2022AutoSDF}
Paritosh Mittal, Y. Cheng, Maneesh Singh, and Shubham Tulsiani.
\newblock Autosdf: Shape priors for 3d completion, reconstruction and
  generation.
\newblock 2022.

\bibitem{Mo2019PartNet}
Kaichun Mo, Shilin Zhu, Angel~X. Chang, L. Yi, Subarna Tripathi, L. Guibas, and
  H. Su.
\newblock Partnet: A large-scale benchmark for fine-grained and hierarchical
  part-level 3d object understanding.
\newblock {\em 2019 IEEE/CVF Conference on Computer Vision and Pattern
  Recognition (CVPR)}, pages 909--918, 2019.

\bibitem{nash2020polygen}
Charlie Nash, Yaroslav Ganin, SM~Ali Eslami, and Peter Battaglia.
\newblock Polygen: An autoregressive generative model of 3d meshes.
\newblock In {\em International conference on machine learning}, pages
  7220--7229. PMLR, 2020.

\bibitem{nash2021DCTransformer}
Charlie Nash, Jacob Menick, Sander Dieleman, and Peter~W Battaglia.
\newblock Generating images with sparse representations.
\newblock {\em arXiv preprint arXiv:2103.03841}, 2021.

\bibitem{Niemeyer2019oflow}
M. Niemeyer, Lars~M. Mescheder, Michael Oechsle, and Andreas Geiger.
\newblock Occupancy flow: 4d reconstruction by learning particle dynamics.
\newblock {\em ICCV}, pages 5378--5388, 2019.

\bibitem{oord2016pixelcnn}
A{\"a}ron van~den Oord, Nal Kalchbrenner, Oriol Vinyals, Lasse Espeholt, Alex
  Graves, and Koray Kavukcuoglu.
\newblock Conditional image generation with pixelcnn decoders.
\newblock In {\em Proc. Conf. on Neural Information Processing Systems}, pages
  4797--4805, 2016.

\bibitem{park2019deepsdf}
Jeong~Joon Park, Peter Florence, Julian Straub, Richard Newcombe, and Steven
  Lovegrove.
\newblock Deepsdf: Learning continuous signed distance functions for shape
  representation.
\newblock In {\em Proc. IEEE Conf. on Computer Vision \& Pattern Recognition},
  pages 165--174, 2019.

\bibitem{parmar2018imagetransformer}
Niki Parmar, Ashish Vaswani, Jakob Uszkoreit, Lukasz Kaiser, Noam Shazeer,
  Alexander Ku, and Dustin Tran.
\newblock Image transformer.
\newblock In {\em International Conference on Machine Learning}, pages
  4055--4064. PMLR, 2018.

\bibitem{Peng2020ConvONet}
Songyou Peng, Michael Niemeyer, Lars Mescheder, Marc Pollefeys, and Andreas
  Geiger.
\newblock Convolutional occupancy networks.
\newblock In {\em Proc. Euro. Conf. on Computer Vision}, 2020.

\bibitem{radford2019GPT2}
Alec Radford, Jeff Wu, Rewon Child, David Luan, Dario Amodei, and Ilya
  Sutskever.
\newblock Language models are unsupervised multitask learners.
\newblock 2019.

\bibitem{ramesh2021dalle}
Aditya Ramesh, Mikhail Pavlov, Gabriel Goh, Scott Gray, Chelsea Voss, Alec
  Radford, Mark Chen, and Ilya Sutskever.
\newblock Zero-shot text-to-image generation.
\newblock {\em arXiv preprint arXiv:2102.12092}, 2021.

\bibitem{razavi2019vqvae2}
Ali Razavi, Aaron van~den Oord, and Oriol Vinyals.
\newblock Generating diverse high-fidelity images with vq-vae-2.
\newblock {\em arXiv preprint arXiv:1906.00446}, 2019.

\bibitem{rock2015completing}
Jason Rock, Tanmay Gupta, Justin Thorsen, JunYoung Gwak, Daeyun Shin, and Derek
  Hoiem.
\newblock Completing 3d object shape from one depth image.
\newblock In {\em Proceedings of the IEEE Conference on Computer Vision and
  Pattern Recognition}, pages 2484--2493, 2015.

\bibitem{shu2019treegan}
Dong~Wook Shu, Sung~Woo Park, and Junseok Kwon.
\newblock 3d point cloud generative adversarial network based on tree
  structured graph convolutions, 2019.

\bibitem{stutz2018completion}
David Stutz and Andreas Geiger.
\newblock Learning 3d shape completion from laser scan data with weak
  supervision.
\newblock In {\em Proceedings of the IEEE Conference on Computer Vision and
  Pattern Recognition (CVPR)}, June 2018.

\bibitem{sun2020pointgrow}
Yongbin Sun, Yue Wang, Ziwei Liu, Joshua Siegel, and Sanjay Sarma.
\newblock Pointgrow: Autoregressively learned point cloud generation with
  self-attention.
\newblock In {\em Proceedings of the IEEE/CVF Winter Conference on Applications
  of Computer Vision}, pages 61--70, 2020.

\bibitem{tatarchenko2019single}
Maxim Tatarchenko, Stephan~R Richter, Ren{\'e} Ranftl, Zhuwen Li, Vladlen
  Koltun, and Thomas Brox.
\newblock What do single-view {3D} reconstruction networks learn?
\newblock In {\em Proc. IEEE Conf. on Computer Vision \& Pattern Recognition},
  pages 3405--3414, 2019.

\bibitem{Tchapmi2019topnet}
Lyne~P. Tchapmi, Vineet Kosaraju, Hamid Rezatofighi, Ian Reid, and Silvio
  Savarese.
\newblock Topnet: Structural point cloud decoder.
\newblock In {\em Proc. IEEE Conf. on Computer Vision \& Pattern Recognition},
  2019.

\bibitem{uria2016nade}
Benigno Uria, Marc-Alexandre C{\^o}t{\'e}, Karol Gregor, Iain Murray, and Hugo
  Larochelle.
\newblock Neural autoregressive distribution estimation.
\newblock {\em The Journal of Machine Learning Research}, 17(1):7184--7220,
  2016.

\bibitem{van2017vqvae1}
A{\"a}ron van~den Oord, Oriol Vinyals, and Koray Kavukcuoglu.
\newblock Neural discrete representation learning.
\newblock In {\em Proc. Conf. on Neural Information Processing Systems}, 2017.

\bibitem{van2016pixelrnn}
Aaron Van~Oord, Nal Kalchbrenner, and Koray Kavukcuoglu.
\newblock Pixel recurrent neural networks.
\newblock In {\em International Conference on Machine Learning}, pages
  1747--1756. PMLR, 2016.

\bibitem{vaswani2017transformer}
Ashish Vaswani, Noam Shazeer, Niki Parmar, Jakob Uszkoreit, Llion Jones,
  Aidan~N Gomez, Lukasz Kaiser, and Illia Polosukhin.
\newblock Attention is all you need.
\newblock In {\em NIPS}, 2017.

\bibitem{wan2021ict}
Ziyu Wan, Jingbo Zhang, Dongdong Chen, and Jing Liao.
\newblock High-fidelity pluralistic image completion with transformers.
\newblock {\em arXiv preprint arXiv:2103.14031}, 2021.

\bibitem{wang2018pixel2mesh}
Nanyang Wang, Yinda Zhang, Zhuwen Li, Yanwei Fu, Wei Liu, and Yu-Gang Jiang.
\newblock Pixel2mesh: Generating 3d mesh models from single rgb images, 2018.

\bibitem{wang2020sceneformer}
Xinpeng Wang, Chandan Yeshwanth, and Matthias Nie{\ss}ner.
\newblock Sceneformer: Indoor scene generation with transformers.
\newblock {\em arXiv preprint arXiv:2012.09793}, 2020.

\bibitem{wu2020cGAN}
Rundi Wu, Xuelin Chen, Yixin Zhuang, and Baoquan Chen.
\newblock Multimodal shape completion via conditional generative adversarial
  networks.
\newblock In {\em Proc. Euro. Conf. on Computer Vision}, August 2020.

\bibitem{Xiang2021SnowflakeNet}
Peng Xiang, Xin Wen, Yu-Shen Liu, Yan-Pei Cao, Pengfei Wan, Wen Zheng, and
  Zhizhong Han.
\newblock Snowflakenet: Point cloud completion by snowflake point deconvolution
  with skip-transformer.
\newblock {\em 2021 IEEE/CVF International Conference on Computer Vision
  (ICCV)}, pages 5479--5489, 2021.

\bibitem{yu2021pointr}
Xumin Yu, Yongming Rao, Ziyi Wang, Zuyan Liu, Jiwen Lu, and Jie Zhou.
\newblock Pointr: Diverse point cloud completion with geometry-aware
  transformers.
\newblock In {\em ICCV}, 2021.

\bibitem{Yu2021PointBERT}
Xumin Yu, Lulu Tang, Yongming Rao, Tiejun Huang, Jie Zhou, and Jiwen Lu.
\newblock Point-bert: Pre-training 3d point cloud transformers with masked
  point modeling.
\newblock {\em ArXiv}, abs/2111.14819, 2021.

\bibitem{yuan2018pcn}
Wentao Yuan, Tejas Khot, David Held, Christoph Mertz, and Martial Hebert.
\newblock Pcn: Point completion network.
\newblock In {\em 2018 International Conference on 3D Vision (3DV)}, pages
  728--737, 2018.

\bibitem{zhang2021shapeinversion}
Junzhe Zhang, Xinyi Chen, Zhongang Cai, Liang Pan, Haiyu Zhao, Shuai Yi,
  Chai~Kiat Yeo, Bo Dai, and Chen~Change Loy.
\newblock Unsupervised 3d shape completion through gan inversion.
\newblock In {\em CVPR}, 2021.

\bibitem{zhou2021pvd}
Linqi Zhou, Yilun Du, and Jiajun Wu.
\newblock 3d shape generation and completion through point-voxel diffusion.
\newblock In {\em Proc. Int. Conf. on Computer Vision}, pages 5826--5835,
  October 2021.

\end{thebibliography}
